\documentclass[pdflatex,sn-mathphys-num,iicol]{sn-jnl}

\usepackage[table]{xcolor}
\usepackage{graphicx}%
\usepackage{multirow}%
\usepackage{amsmath,amssymb,amsfonts}%
\usepackage{amsthm}%
\usepackage{mathrsfs}%
\usepackage{textcomp}%
\usepackage{manyfoot}%
\usepackage{booktabs}%
\usepackage{algorithm}%
\usepackage{algorithmicx}%
\usepackage{algpseudocode}%
\usepackage{listings}%
\usepackage[export]{adjustbox}
\makeatletter
\DeclareRobustCommand\onedot{\futurelet\@let@token\@onedot}
\def\@onedot{\ifx\@let@token.\else.\null\fi\xspace}

\def\eg{\emph{e.g}\onedot} 
\def\ie{\emph{i.e}\onedot} 
\def\cf{\emph{cf}\onedot} 
\def\etc{\emph{etc}\onedot} 
\def\wrt{w.r.t\onedot} 
 
\def\etal{\emph{et al}\onedot}
\makeatother

\newcommand{\newtext}[1]{\textcolor{red}{#1}}

\newcommand{\newnewtext}[1]{\textcolor{purple}{#1}}

\theoremstyle{thmstyleone}%

\theoremstyle{thmstyletwo}%

\theoremstyle{thmstylethree}%

\usepackage{graphicx}
\usepackage{booktabs}
\usepackage{comment}
\usepackage{amsmath,amssymb}
\usepackage{color}
\usepackage{url}
\usepackage{booktabs}
\usepackage{pifont}
\usepackage{multirow}
\usepackage{xspace}
\usepackage{multicol}
\usepackage{makecell}
\usepackage{subcaption}
\usepackage{array}
\usepackage{orcidlink}
\usepackage[accsupp]{axessibility}  %

\usepackage{letltxmacro}
\usepackage{xparse}

\usepackage{pgfplots}
\pgfplotsset{width=10cm,compat=1.9}

\definecolor{Seaborn1}{HTML}{1f77b4}
\definecolor{Seaborn2}{HTML}{ff7f0e}
\definecolor{Seaborn3}{HTML}{2ca02c}
\definecolor{Seaborn4}{HTML}{d62728}
\definecolor{Seaborn5}{HTML}{9467bd}
\definecolor{Seaborn6}{HTML}{8c564b}
\definecolor{Seaborn7}{HTML}{e377c2}
\definecolor{Seaborn8}{HTML}{7f7f7f}
\definecolor{Seaborn9}{HTML}{bcbd22}
\definecolor{Seaborn10}{HTML}{17becf}

\definecolor{best}{rgb}{1.0, 0.6, 0.6}    %
\definecolor{second}{rgb}{1.0, 0.8, 0.6}  %

\LetLtxMacro{\OriginalCellColor}{\cellcolor}

\RenewDocumentCommand\cellcolor{m m}{%
  \OriginalCellColor{#1}%
  #2%
}

\newcommand{\bestword}[1]{\colorbox{best}{#1}}
\newcommand{\secondword}[1]{\colorbox{second}{#1}}

\newcommand{\M}[1]{\mathbf{#1}}
\newcommand{\dataset}[1]{{\fontfamily{cmtt}\selectfont #1}\xspace }

\newcommand{\ETH}{\dataset{ETH3D}}
\newcommand{\ROTUNDA}{\dataset{ROTUNDA}}
\newcommand{\VITUS}{\dataset{CATHEDRAL}}
\newcommand{\CATHEDRAL}{\dataset{CATHEDRAL}}

\newcommand{\PP}{\dataset{PragueParks}}
\newcommand{\Euroc}{\dataset{EuRoC-MAV}}

\newcommand{\ZK}[1]{{\textcolor{cyan}{ZK: {[#1]}}}}

\newcommand{\vk}[1]{{\textcolor{magenta}{VK: {[#1]}}}}
\newcommand{\F}{{$\M F$}}
\newcommand{\E}{{$\M E$}}
\newcommand{\Fk}{{$\M F \lambda$}}
\newcommand{\Fkk}{{$\M F \lambda_1\lambda_2$}}
\def\gb{Gr{\"o}bner basis\xspace}

\newcommand{\R}{\M R}
\newcommand{\tvec}{\vec{t}}
\newcommand{\Ef}{$\M Ef$}
\newcommand{\g}{\vec{g}}

\renewcommand{\vk}[1]{#1}

\renewcommand{\ZK}[1]{#1}
\renewcommand{\newtext}[1]{#1}
\renewcommand{\newnewtext}[1]{#1}

\begin{document}

\title[Article Title]{Are Minimal Radial Distortion Solvers Really Necessary for Relative Pose Estimation?}

\author[1]{\fnm{Viktor} \sur{Kocur}\,\orcidlink{https://orcid.org/0000-0001-8752-2685}}\email{viktor.kocur@fmph.uniba.sk}
\equalcont{These authors contributed equally to this work.}

\author*[2]{\fnm{Charalambos} \sur{Tzamos}\,\orcidlink{https://orcid.org/0000-0001-7029-6431}}\email{tzamocha@fel.cvut.cz}
\equalcont{These authors contributed equally to this work.}

\author[2]{\fnm{Yaqing} \sur{Ding}\,\orcidlink{https://orcid.org/0000-0002-7448-6686}}\email{yaqing.ding@fel.cvut.cz}

\author[1]{\fnm{Zuzana} \sur{Berger Haladova}\,\orcidlink{https://orcid.org/0000-0002-5947-8063}}\email{haladova@fmph.uniba.sk}

\author[3]{\fnm{Torsten} \sur{Sattler}\,\orcidlink{https://orcid.org/0000-0001-9760-4553}}\email{torsten.sattler@cvut.cz}

\author[2]{\fnm{Zuzana} \sur{Kukelova}\,\orcidlink{https://orcid.org/0000-0002-1916-8829}}\email{kukelzuz@fel.cvut.cz}

\affil[1]{\orgdiv{Faculty of Mathematics, Physics and Informatics}, \orgname{Comenius University Bratislava}, \orgaddress{\city{Bratislava}, \country{Slovakia}}}

\affil[2]{\orgdiv{Visual Recognition Group}, \orgname{Faculty of Electrical Engineering, Czech Technical University in Prague}, \orgaddress{\city{Prague}, \country{Czech Republic}}}

\affil[3]{\orgdiv{Czech Institute of Informatics, Robotics and Cybernetics}, \orgname{Czech Technical University in Prague}, \orgaddress{\city{Prague}, \country{Czech Republic}}}

\abstract{Estimating the relative pose between two cameras is a fundamental step in many applications such as Structure-from-Motion. The common approach to relative pose estimation is to apply a minimal solver inside a RANSAC loop. Highly efficient solvers exist for pinhole cameras. Yet, (nearly) all cameras exhibit radial distortion. Not modeling radial distortion leads to (significantly) worse results. However, minimal radial distortion solvers are significantly more complex than pinhole solvers, both in terms of run-time and implementation efforts. This paper compares radial distortion solvers with \newnewtext{two} simple-to-implement approaches \newnewtext{that do not use minimal radial distortion solvers: The first approach}  combines an efficient pinhole solver with sampled radial undistortion parameters,  
\newnewtext{where the sampled parameters are used for undistortion prior to applying the pinhole solver. The second approach uses a state-of-the-art neural network to estimate the distortion parameters rather than sampling them from a set of potential values}. 
Extensive experiments on multiple datasets, 
\newtext{and different camera setups,} show that 
\newtext{complex minimal radial distortion solvers are not necessary in practice. We discuss under which conditions a simple sampling of radial undistortion parameters is preferable over calibrating cameras using a learning-based prior approach.}
Code and \newnewtext{newly created} benchmark \newnewtext{for relative pose estimation under radial distortion} are available at {\url{https://github.com/kocurvik/rdnet}}.     
}

\maketitle

\section{Introduction}
\label{sec:introduction}
\noindent Estimating the relative pose of two cameras, \ie, estimating the relative rotation, translation, and potentially internal calibration parameters of both cameras, is a fundamental problem in computer vision. %
Relative pose solvers are core components of Structure-from-Motion (SfM)~\cite{wu2011visualsfm,Schoenberger2016CVPR} and localization pipelines~\cite{sattler2016efficient,svarm2016city,sarlin2019coarse} and play an important role in robotics~\cite{mur2015orb,mur2017orb} and autonomous driving~\cite{scaramuzza2011visual}.

A predominant way to estimate the relative pose of two cameras %
is based on 2D-2D point correspondences between the two images. Due to noise and the presence of outliers, %
robust estimation algorithms, such as RANdom SAmple Consensus (RANSAC)~\cite{Fischler-Bolles-ACM-1981}, or its more modern variants~\cite{raguram2013usac,barath2017graph}, are used for the estimation.
Inside RANSAC two different steps are performed: 
(i) Estimating the camera geometry from a (small or minimal) sample of correspondences and classifying all correspondences into inliers and outliers \wrt the obtained camera model.
(ii) Local optimization (LO) of the camera model parameters %
on (a subset of) the inliers to better account for noise in the 2D point positions~\cite{Chum-2003,lebeda2012fixing}.

The main objective of the first step is to obtain a  %
camera geometry estimate and the subset of correspondences  consistent with it. %
Small samples %
are preferable since the %
number of RANSAC iterations, and thus the run-time, depends exponentially on the number of correspondences required for model estimation. 
Solvers that estimate the camera geometry using a minimal number of correspondences and using all available polynomial constraints are known as minimal solvers.
The most commonly used minimal solvers for relative pose estimation 
are the well-known 5-point solver~\cite{nister2004efficient} 
for calibrated cameras and the 7-point solver~\cite{hartley2003multiple} for uncalibrated cameras. 
Both %
are highly efficient. %

Minimal solvers produce estimates that perfectly fit the correspondences in the minimal sample. 
In practice, the 2D point correspondences are noisy and the noise in the 2D coordinates propagates to the estimates. %
The goal of the second step, \ie, LO inside RANSAC, is to reduce the impact of measurement noise on pose accuracy~\cite{lebeda2012fixing}. 
Commonly, a non-minimal solver that fits model parameters to a larger-than-minimal sample is used~\cite{Chum-2003}, or a robust cost function that optimizes model parameters on all inliers is minimized~\cite{lebeda2012fixing}.
The most common %
non-minimal solver for relative pose estimation is the linear 8-point solver~\cite{hartley2003multiple}. 

All previously mentioned solvers, \ie, the 5-point, 7-point, and 8-point solvers, are widely used in SfM pipelines and other applications. 
They are based on the pinhole camera model. %
Yet, virtually all cameras exhibit some amount of radial distortion. %
Ignoring the distortion,
even for standard consumer cameras, can lead to 
errors in 3D reconstruction~\cite{fitzgibbon2001simultaneous}, camera calibration accuracy, \etc %

There are several ways to deal with radial distortion: (1) Ignore radial distortion estimation during RANSAC and model %
it only in a post-processing step, \eg, during bundle adjustment in SfM~\cite{snavely2008modeling,schonberger2016structure}. 
(2) Ignore radial distortion in the first RANSAC step but take it into account in the second step 
(LO), \eg,  %
by using a non-minimal solver that estimates radial undistortion parameters or by modeling distortion %
when minimizing a robust cost function. 
(3) Already estimate the radial distortion in the first RANSAC step (and refine it during LO). %
Approach (3) is the most principled solution as it enables taking radial distortion into account during inlier counting.  
Ignoring radial distortion inside the solver typically leads to identifying only the subset of the inliers that is less affected by the distortion.\footnote{A point in one image maps to an epipolar \textit{curve} in a radially distorted second image. Ignoring radial distortion thus means approximating this curve by a line. %
The approximation is %
only 
good around the center of %
strongly distorted images.} %
As a result of only containing points that are only mildly affected by distortion, this inlier set often does not contain enough information to accurately estimate the undistortion parameters. 
Thus, %
approaches (1) and (2), which operate on the inliers identified beforehand,  %
are likely to fall into local minima, without recovering
 correct distortion and camera parameters. %

Radial distortion modeling is a mathematically challenging task, and even the simplest one-parameter radial distortion model leads to complex polynomial equations when incorporated into relative pose solvers~\cite{fitzgibbon2001simultaneous, kukelova2007minimal}. Thus, algorithms for estimating
epipolar geometry for cameras with radial distortion started appearing only after introducing efficient algebraic polynomial solvers into the computer vision community~\cite{fitzgibbon2001simultaneous,barreto2005fundamental,jin2008three,byrod2009minimal,kukelova2010fast}. 
With improvements in methods for generating efficient polynomial solvers, also minimal radial distortion solvers are improving their efficiency and stability. 
However, compared to solvers for the pinhole camera model, most of these solvers are still orders of magnitude slower,  \eg, 
 the fastest 9-point solver for different distortions runs 210$\mu s$~\cite{Oskarsson_2021_CVPR}, and the 6-point solver with unknown common radial distortion for calibrated cameras runs 1.18$ms$. 
This is significantly slower than the 5-point and 7-point pinhole camera solvers that run in less than 6$\mu s$. 
Moreover, since these solvers estimate more unknown parameters, they need to sample more points inside RANSAC.\footnote{Usually one more point for cameras with a common unknown radial distortion and two more points for cameras with different radial distortions.} They also return more potential solutions to the camera model.
Thus, even though radial distortion solvers estimate models that better fit the data, they may require more RANSAC iterations and longer RANSAC run-times.  

Radial distortion solvers are not only slower but also more complex to implement. %
At the same time, many of the existing minimal radial distortion solvers do not have a publicly available implementation. %
Even though the papers that present novel radial distortion solvers show advantages of these solvers on real data, they usually focus on presenting novel parameterizations and solution strategies and their numerical stability on synthetic data. Real experiments are mostly limited to small datasets (\eg, a single scene), simpler variants of RANSAC,  and qualitative instead of quantitative results.   
The above-mentioned facts are most likely %
the reasons why (minimal) radial distortion solvers are not often used in practice. Instead, it is common to use either approach (1) or (2)~\cite{schonberger2016structure,snavely2008modeling}. %
Naturally, this leads to the question whether 
(minimal) radial distortion solvers are actually necessary %
in practical applications. %

The goal of this paper is to answer this question. 
To this end, 
\newnewtext{we introduce two new approaches to relative pose estimation under radial distortion: (1)} We introduce a new sampling-based strategy 
\newtext{that combines an efficient solver for uncalibrated relative pose problem (e.g. the 7-point \F \ solver~\cite{hartley2003multiple}, or the 6-point \Ef \ solver~\cite{larsson2017efficient})} with a sampled undistortion parameter:
In each RANSAC iteration, we run the solver potentially several times (1-3x) with different (but fixed) undistortion parameters. \newtext{
\newnewtext{(2) Rather than sampling from a fixed set of undistortion parameters, the second approach uses a learning-based prior for the parameters: 
A neural network~\cite{veicht2024geocalib} predicts the radial undistortion (and potentially other camera) parameters, which are then used to obtain initial pose estimates inside RANSAC.}
}
The paper makes the following contributions:
\begin{enumerate}
\item We extensively evaluate different approaches for \newtext{uncalibrated} relative pose estimation under radial distortion on multiple datasets, under different scenarios.
We are not aware of any such practical evaluation of radial distortion solvers in the literature. 
\item \newtext{We show that both the sampling-based and learning-based prior strategies, which are both easy to implement, perform similar or better than most accurate radial distortion solvers.} We thus show that dedicated minimal radial distortion solvers for the relative pose problem are not necessary in practice.  
\newtext{\item We show that for the case of two cameras with unknown and shared intrinsics, the sampling-based strategy combined with the 6-point \Ef \ solver~\cite{larsson2017efficient} provides similar accuracy to the learning-based prior approach while its total runtime is lower and it does not require a GPU to run efficiently.}
\newtext{\item For the case of two cameras with different intrinsics, we show that the simple sampling-based strategy performs better than other methods when the time budget for computation is limited ($<$100 ms) or the computation has to be performed on low-cost hardware.}
\item{\ZK{We show that the sampling-based strategy is more robust and works for all types of data compared to learning-based priors that can be less precise for some types of data (\eg images outside of the training distribution).}}
\item We create a new benchmark, consisting of two scenes, containing images taken with different cameras with multiple different distortions. 
\item Code and dataset are available at \url{https://github.com/kocurvik/rdnet}. 
\end{enumerate}

\newnewtext{This paper is an extension of our previous work~\cite{Tzamos2024ECCVW}. 
Compared to~\cite{Tzamos2024ECCVW}, 
this work strengthen the main message of the original work, namely that minimal radial distortion solvers are not necessary, by 
(1) including a new approach to uncalibrated relative pose estimation under radial distortion that does not rely on a minimal radial distortion solver (the learning-based prior strategy introduced in Sec.~\ref{sec:learning-based_strategy}); 
(2) performing experiments on two additional datasets (PragueParks~\cite{IMC2020} and EuRoc-MaV~\cite{Burri25012016}) from the literature, significantly increasing the number of image pairs that are used for evaluation by 13.7k pairs; 
(3) evaluating additional minimal solvers for relative pose estimation together with our two strategies (sampling-based and learning-based prior-based approaches); 
and (4) providing insights into which of our strategies is preferable under which conditions.}

\section{Related Work}
The literature studies three groups of radial distortion relative pose problems: Two cameras with equal and unknown radial distortion; 
two cameras where only one has unknown %
distortion; %
two cameras with different and unknown distortion.

\subsection{Equal and unknown radial distortion} Fitzgibbon~\cite{fitzgibbon2001simultaneous} 
introduced an one-parameter division model for modeling undistortion and an algorithm for estimating the fundamental matrix with equal and unknown radial distortion using this model.
This algorithm does not use %
the singularity constraint on the fundamental matrix, necessitating 9 point correspondences instead of the minimal 8. 
This approach 
transforms the problem into a standard quadratic eigenvalue problem with %
up to 10 solutions. 
The first minimal solution for epipolar geometry estimation with %
the one-parameter division model using 8 point correspondences %
was proposed by~\cite{kukelova2007minimal}, 
using the \gb method~\cite{cox2005using} to solve a system of polynomial equations. This solver has been improved by %
using an automatic generator of
\gb solvers~\cite{kukelova2008automatic}, performing Gauss-Jordan (G-J)
elimination of a 32×48 matrix and eigenvalue decomposition of
a 16×16 matrix, which has up to 16 solutions. 
Jiang~\etal~\cite{jiang2015minimal}, used 7 point correspondences to solve the problem of essential matrix estimation for two cameras with equal and unknown focal length and radial distortion. 
This problem results in a complex system of polynomial equations and a large solver that performs the LU decomposition of an 886×1011 matrix and computes the eigenvalues of a 68×68 matrix. 
Thus, this solver is too %
time-consuming %
for practical %
applications. A similar but more efficient solver was proposed by Oskarsson~\cite{Oskarsson_2021_CVPR}, however the solver is highly unstable making it impractical.

\subsection{One unknown radial distortion} 
Kuang~\etal~\cite{kuang2014minimal} studied three minimal cases for relative pose estimation with a
single unknown radial distortion based on the \gb method: 8-point fundamental matrix and radial distortion; 7-point essential matrix, focal length and radial distortion; 6-point essential matrix and radial distortion. However, these solvers assume one of the two cameras has known or no radial distortion. In many scenarios, this assumption does not hold.

\subsection{Different and unknown radial distortions}
All of the above mentioned algorithms estimate only one radial undistortion parameter for one or both cameras. 
In practice, \eg, using images %
downloaded from the Internet, two cameras can %
have different and unknown radial distortions. The problem of fundamental matrix estimation with different and unknown radial distortions, $\M F \lambda_1 \lambda_2$, was first studied by Barreto and Daniilidis~\cite{barreto2005fundamental}, 
proposing %
a non-minimal linear algorithm %
using 15 point correspondences (F15). %
The minimal 9-point case (F9) for this problem was studied in~\cite{kukelova2007two,byrod2008fast,kukelova2008automatic,kukelova2010fast}. 
The solver from Byr\"od~\etal~\cite{byrod2008fast} (F9) %
performs LU decomposition of a 393×389 matrix, SVD decomposition of a 69×69 matrix, and %
eigenvalue computation of a 24×24 matrix. 
In~\cite{kukelova2008automatic}, a faster version based on a \gb ($\rm F9_{\rm A}$) was proposed. 
It performs G-J elimination of a 179×203 matrix and eigenvalue decomposition of a 24×24 matrix. 
However, this solver is slightly less stable than F9, and still too slow for real-time applications. 
Kukelova~\etal~\cite{kukelova2010fast} suggested an efficient, non-minimal solver using 12 point correspondences (F12) that %
generates up to four real solutions. %
However, %
this algorithm %
is more sensitive to noise than the minimal %
$\rm F9_{\rm A}$. %
Balancing efficiency and  noise sensitivity, ~\cite{kukelova2015efficient} proposed a 10-point solver %
that is much faster than the minimal 9-point solver and more robust to image noise than %
the 12-point solver.

Recently,~\cite{Oskarsson_2021_CVPR} presented a unified formulation for relative pose problems involving radial distortion and proposed more efficient minimal solvers for all different configurations. 
While some of the proposed solvers are already quite efficient, \eg, the 8-point solver for uncalibrated cameras with common radial distortion, others, like the 9-point solver for different distortions, are still too slow and/or numerically unstable to be useful in practice.

\subsection{Parameter sampling}
Instead of jointly estimating the absolute camera pose and the focal length of an uncalibrated camera, Sattler~\etal~\cite{Sattler2014ECCV} proposed a RANSAC variant that combines parameter sampling and parameter estimation. 
In each RANSAC iteration, they first randomly sample a focal length value and then estimate the pose of the now-calibrated camera. 
The probability distribution over the focal length values is then updated based on the number of inliers of the estimated pose. 
We propose a simpler sampling-based strategy for relative pose estimation that uses a small fixed set of undistortion parameters.  %
In contrast to~\cite{Sattler2014ECCV}, our approach can  easily be applied to 2D sampling problems, \eg, two different and unknown undistortion parameters.

\section{Radial Distortion \newnewtext{Estimation}}%
\label{sec:rd_solvers}
\subsection{Background}
A pair of corresponding distorted image points $\mathbf{x}_i \leftrightarrow \mathbf{x}_i^\prime$,  detected in two uncalibrated images, is related by the epipolar constraint
\begin{equation}
    u(\mathbf{x}_i, \mathbf{\Lambda})^\top \M F u(\mathbf{x}_i^\prime, \mathbf{\Lambda}^\prime) = 0 \enspace ,\label{eq:01}
\end{equation}
where $\mathbf{x}_i, \mathbf{x}_i^\prime \in \mathbb{P}^2$, $\M F$ is the fundamental matrix encoding the relative pose and the internal calibrations of the two cameras, and 
$u:\mathbb{P}^2 \times \mathbb{R}^{n} \rightarrow \mathbb{P}^2$ denotes an undistortion function, which is a function of the distorted image point $\mathbf{x}_i$ and $n$ undistortion parameters $\mathbf{\Lambda} \in \mathbb{R}^{n}$.

In this paper, we model the undistortion function using the one-parameter division model~\cite{fitzgibbon2001simultaneous}. 
In this model, given an observed radially distorted point with homogeneous coordinates $\mathbf{x} = [x_d, y_d, 1]^\top$, and the undistortion parameter $\lambda \in \mathbb{R}$, the %
undistorted image point is given as %
\begin{equation}
    u(\mathbf{x}, \lambda) = [x_d, y_d, 1 + \lambda (x_d^2 + y_d^2)]^\top \enspace ,
    \label{eq:02}
\end{equation}
\noindent assuming that the distortion center is in the image center. This model is very commonly used in practice due to its simplicity, efficiency, and robustness, since it can adequately capture %
even large distortions of wide-angle lenses. It is incorporated in almost all minimal and non-minimal radial distortion solvers.

\subsection{Radial Distortion Solvers}
\label{seq:solvers:rad_dist_solvers}
The goal of this paper is not to introduce novel minimal or non-minimal radial distortion solvers, but to study the performance of the existing solvers under different conditions. We study the two most practical scenarios of two uncalibrated cameras with unknown (i) equal and (ii) different radial distortions. 
We denote these problems as (i) the \Fk \ and (ii) the \Fkk \ problems.\footnote{Instead of $\lambda$ and $\lambda^{\prime}$ as used in~\eqref{eq:01}, we use $\lambda_1$ and $\lambda_2$ for better readability.} Next, we briefly describe the radial distortion solvers for these two problems studied %
in this paper, as well as some improvements to these solvers.

\noindent \Fk \ : Assuming equal unknown distortion modeled using the one-parameter division model~\eqref{eq:02}, the relative pose problem for uncalibrated cameras has 8 degrees of freedom (DoF).  For this problem, we consider the following solvers: 
\begin{itemize}
    \item 8pt \Fk \ : Among all minimal 8pt solvers~\cite{kukelova2007minimal,kukelova2008automatic,larsson2017efficient,Oskarsson_2021_CVPR}, the solver from~\cite{Oskarsson_2021_CVPR} is the most efficient. It formulates the elements of $\M{F}$ as functions of the undistortion parameter $\lambda$, and obtains an univariate polynomial in $\lambda$ of degree 16, which can be solved using the Sturm sequences, with up to 16 solutions. 

    \item 9pt \Fk \ : By ignoring the $\det(\M{F}) = 0$ constraint, the \Fk \ 
 problem can be solved using nine point correspondences. Fitzgibbon~\cite{fitzgibbon2001simultaneous} solves the nine equations~\eqref{eq:01} by converting them into a polynomial eigenvalue problem.
 While~\cite{fitzgibbon2001simultaneous} was able to remove several spurious solutions by transforming the original eigenvalue problem of size $18 \times 18$ into a problem of size $10 \times 10$,~\cite{fitzgibbon2001simultaneous} also observed that 4 of the 10 solutions of this system are imaginary. In this paper, we propose a modification of the solver proposed in~\cite{fitzgibbon2001simultaneous}, in which we directly remove 4 imaginary solutions, resulting in a more efficient solver that needs to find the eigenvalues of a smaller $6\times 6$ matrix. To remove these 4 imaginary solutions, we use the structure of matrices that appear in the polynomial eigenvalue formulation of this problem and the method proposed in~\cite{kukelova2012polynomial}. For more details, see Sec.~\ref{sec:ref_solvers_equal}.
\end{itemize}

\noindent \Fkk \ : For the case of %
different unknown radial distortions, we have 9 DoF. For this problem, we consider the following solvers:
\begin{itemize}
    \item  9pt \Fkk \ : Equations for cameras with different unknown distortions are more complex than for the equal distortion case. 
    In this case the system of equations has 24 solutions and the fastest Gr\"{o}bner basis solver from~\cite{larsson2017efficient}, which returns 24 solutions, performs elimination of a large matrix of size $84 \times 117$ followed by the eigendecomposition of a $24 \times 24$ matrix. The recently published parameterization of this problem in~\cite{Oskarsson_2021_CVPR} performs elimination of a smaller matrix of size $51 \times 99$ followed by the  eigendecomposition of a $48 \times 48$ matrix. The solver returns up to 48 solutions. 
    However, it is still faster than the solver from~\cite{larsson2017efficient}. Thus, in our experiments, we use the solver from~\cite{Oskarsson_2021_CVPR}. 
    \item  10pt \Fkk \ : In~\cite{kukelova2015efficient} it was shown that in many scenarios inside RANSAC it is preferable to sample 10 instead of 9 points and run the more efficient 10pt solver. In~\cite{kukelova2015efficient}, the authors proposed several variants of the 10pt solver. In this paper, we use the variant based on a \gb, made available by the authors. The 10pt solvers cannot be easily modified to work with more than 10 points, and thus we use this solver only in the first step of RANSAC, \ie, instead of the minimal solver, and not in the LO step. 
\end{itemize}

\subsection{Modified Solver for Non-minimal Fitting}
\label{sec:refined_solvers}
\ZK{In this section, we describe the proposed modification to the polynomial eigenvalue 9pt \Fk \ solver, in which we remove spurious solutions. We also discuss how to extend these solvers to work with more points.}%

\subsubsection{ \Fk \ solver for equal and unknown distortion}
\label{sec:ref_solvers_equal}
Based on~\cite{fitzgibbon2001simultaneous}, the epipolar constraint with equal and unknown radial distortion can be written as
\begin{equation}
\resizebox{0.89\columnwidth}{!}{$
\begin{matrix}
& & [ & x'_d x_d & x'_d y_d & x'_d & y'_d x_d & y'_d y_d & y'_d & x_d & y_d & 1 & ]& \cdot &  \M f &   \\
+& \lambda & [ & 0 & 0 & x_d r^2 & 0 & 0 & y'_d r^2 & x_d r'^2 & y_d r'^2 & r^2+r'^2 & ]&\cdot &  \M f & \\
+ &\lambda^2 & [ & 0 & 0 &0 & 0 & 0 & 0 & 0 & 0 & r^2 r'^2 &]& \cdot &  \M f & = & 0 \enspace ,
\end{matrix}\label{eq:03}
$}
\end{equation}
where $\M f$ is a $9\times 1$ vector that contains the entries of the fundamental matrix \F\ in row-wise order and $r, r^\prime$ denote the Euclidean distances of the distorted points $\mathbf{x_i}, \mathbf{x_i^\prime}$, respectively, to the center of distortion. It is common to assume that the center of distortion is in the center of the image, \ie, $r = \sqrt{x_d^2 + y_d^2}$. 

\ZK{For n point correspondences,~\eqref{eq:03} can be written in a matrix form}
\begin{equation}
(\M A_0 + \lambda \M A_1 + \lambda^2 \M A_2)\M f = \M 0 \enspace, \label{eq:04}
\end{equation}
\ZK{where $\M A_0,\M A_1 $ and $\M A_2$ are $n \times 9$ coefficient matrices.}
For 9 point correspondences in the 9pt \Fk\ solver, %
equation \eqref{eq:04} defines a polynomial eigenvalue problem that can be solved by computing the eigenvalues of a $18 \times 18$ matrix. In~\cite{fitzgibbon2001simultaneous}, it was shown how the number of solutions of~\eqref{eq:04} can be reduced from 18 to 10 by transforming the problem to an eigenvalue problem of size $10 \times 10$.
However, in~\cite{fitzgibbon2001simultaneous} it was also noted that 4 of these 10 solutions have been found to be imaginary. In our case, we show that the 4 imaginary solutions can be directly removed and we only need to find the eigenvalues of a $6\times 6$ matrix. Since matrix $\M A_2$ is singular while $\M A_0$ is full-rank, we first let $\sigma = 1/\lambda$. Then~\eqref{eq:04} can be written as 
\begin{equation}
(\M A_2 + \sigma \M A_1 + \sigma^2 \M A_0)\M f = \M 0 \enspace .\label{eq:05}
\end{equation}
Solving for $\sigma$ is equivalent to finding the eigenvalues of the following $18 \times 18$ matrix
\begin{equation}
\M B = \begin{bmatrix}
\M 0 & {\M I}\\
-{\M A}_0^{-1}{\M A}_2 & -{\M A}_0^{-1}{\M A}_1 
\end{bmatrix} \enspace .\label{eq:06}
\end{equation}
There are 8 zero columns in $\M A_2$, and 4 zero columns in $\M A_1$. To solve this problem efficiently, we use the technique from~\cite{kukelova2012polynomial}: the zero columns in $-{\M A}_0^{-1}{\M A}_2$ and $-{\M A}_0^{-1}{\M A}_1$ can be removed together with the corresponding rows. In this case, the size of the matrix $\M B$ is reduced to $6\times 6$, and we only need to find the eigenvalues of a $6\times 6$ matrix. Note that in the solver, we directly construct the reduced $6 \times 6$ matrix and avoid computations on the matrix~\eqref{eq:06}

For the non-minimal case, \ie, the case where the number of point correspondences is larger than 9, $-{\M A}_0^{-1}{\M A}_2$ and $-{\M A}_0^{-1}{\M A}_1$ are solved using linear least squares (which can be efficiently solved using the \texttt{ColPivHouseholderQR} function in the Eigen library~\cite{eigenweb}).

\subsection{Sampling Distortions}
\label{sec:rd_solvers:sampling}
While the non-mininmal 9pt \Fk, and 10pt \Fkk \ solvers are reasonably efficient, the minimal 8pt \Fk \ and especially the 9pt \Fkk{} solver are \ZK{ significantly slower than the minimal uncalibrated 7pt pinhole camera solver~\cite{hartley2003multiple} or even a slightly more complex 6pt \Ef \ solver~\cite{larsson2017efficient} for pinhole cameras with common unknown focal length}. Moreover, the minimal radial distortion solvers return more solutions, \ie, 16, 24, or even 48 compared to the 3 solutions of the 7pt solver \ZK{and 15 solutions of the 6pt \Ef \ solver~\cite{larsson2017efficient}}. More solutions lead to reduced efficiency, since within a RANSAC framework each solution has to be evaluated. 
This, together with the fact that the radial distortion solvers sample more points and solve significantly more complex equations,  motivates a common strategy in which in the first step of RANSAC, \ZK{a solver for pinhole camera without radial distortion, usually the standard 7pt solver, is applied},
and the radial distortion is modeled only in the LO step of RANSAC.%

However, as mentioned in Sec.~\ref{sec:introduction}, %
for images with larger distortion, the standard perspective camera model without distortion may not properly model the data and may thus not return a large-enough subset of the true inliers %
and/or an accurate-enough initial pose estimate. 
Yet, small changes in the undistortion parameter $\lambda$ in~\eqref{eq:02}, in general, do not result in large changes in the projection of points into the image. 
For an undistortion parameter $\lambda$ that is reasonably close to the true parameter $\lambda_\text{true}$, %
we can thus expect that 
applying the 7pt solver on 2D point positions that were undistorted using $\lambda$ can %
result in sufficiently-large inlier sets and  sufficiently-accurate initial poses %
that will lead to good estimates in the LO step. 

\newnewtext{The discussions above motivate a simple sampling-based strategy that we propose in this paper:}  
In each iteration of RANSAC, it runs the standard 7-point \F \ solver~\cite{hartley2003multiple} or \newtext{the 6-point \Ef \ solver~\cite{larsson2017efficient}} on image points undistorted with a fixed radial undistortion parameter sampled from an interval of feasible undistortion parameters.
In this approach, we use the facts that the 7pt \F \ and the 6pt \Ef \ solvers are significantly faster than the minimal radial distortion solvers, and return %
fewer solutions that need to be tested inside RANSAC. Thus, even running the 7pt \F \ or the 6pt \Ef \ solvers several times with different fixed undistortion parameters in each RANSAC iteration may lead to a higher efficiency compared to running the 8pt \Fk \ or 9pt \Fkk \ radial distortion solvers.

The \newtext{best choices for the} number $k$ of runs of the 7pt \F \ \newtext{or the 6pt \Ef \ solver} in each iteration, and the values $\M U_i = \{\hat{\lambda}_i^1, \hat{\lambda}_i^2, ..., \hat{\lambda}_i^k\}$, which are used to undistort points in the two cameras $i=1,2$, can differ depending on the application and input data.
In our experiments, we test three variants of the sampling solver: (1) $\M U_1 = \M U_2 = \{ 0 \}$,
which represents the above mentioned standard baseline that assumes no distortion in the first step of RANSAC; 
(2) $\M U_i = \{ \hat{\lambda}_i \}$, $i=1,2$
where we run the 7pt \F \ or the 6pt \Ef \ solver only once for one fixed value of $\hat{\lambda}_i \neq 0$. This can represent a scenario where we have prior knowledge that our images have visible distortion. In our experiments, 
we test a version with $\hat{\lambda}_i$ that represents medium distortion and
can potentially, after LO, provide good results even for cameras with small or large distortion; 
(3)  $\M U_i = \{\hat{\lambda}_i^1, \hat{\lambda}_i^2, \hat{\lambda}_i^3\}$,  where %
we undistort points in each camera with three different fixed parameters that represent, \eg, small, %
medium, %
and large distortion. This setup is, \eg, useful in scenarios where we have images from the ``wild" (\eg, the Internet) that can have a wide variety of different distortions. \ZK{ Note that in this case, if we assume cameras with different distortions, we test only the uncalibrated 7pt \F \ solver, and we run this solver nine times.} Still, this is more efficient than using the %
dedicated distortion \Fkk \ or \Fk \ solvers.

\subsection{Learning-based \newnewtext{Priors for} Radial Distortion Estimation}

\label{sec:learning-based_strategy}
\newnewtext{The solvers discussed in Sections~\ref{seq:solvers:rad_dist_solvers} and~\ref{sec:refined_solvers} estimate the radial undistortion parameter(s) from point correspondences. 
In contrast, our sampling-based approach uses manually selected priors for the undistortion parameters, which can then be refined during LO.} 
\newnewtext{Rather than manually selecting these priors, }
\newtext{%
radial distortion parameters, \newnewtext{as well as other intrinsic parameters such as the focal length,} can also be inferred from a single or multiple images via explicit geometric cues~\cite{lochman2021minimal}, or by using learning-based approaches~\cite{jin2023perspective,veicht2024geocalib}.} %
\newnewtext{To answer the question whether (minimal) radial distortion solvers are necessary in practical applications, %
we thus also evaluate a strategy that uses \ZK{radial distortion and intrinsic priors obtained via learning rather than using manually selected radial distortion priors or estimating radial distortion from point correspondences.}}

\begin{figure}[t!]
    \centering
    \includegraphics[width=0.95\columnwidth]{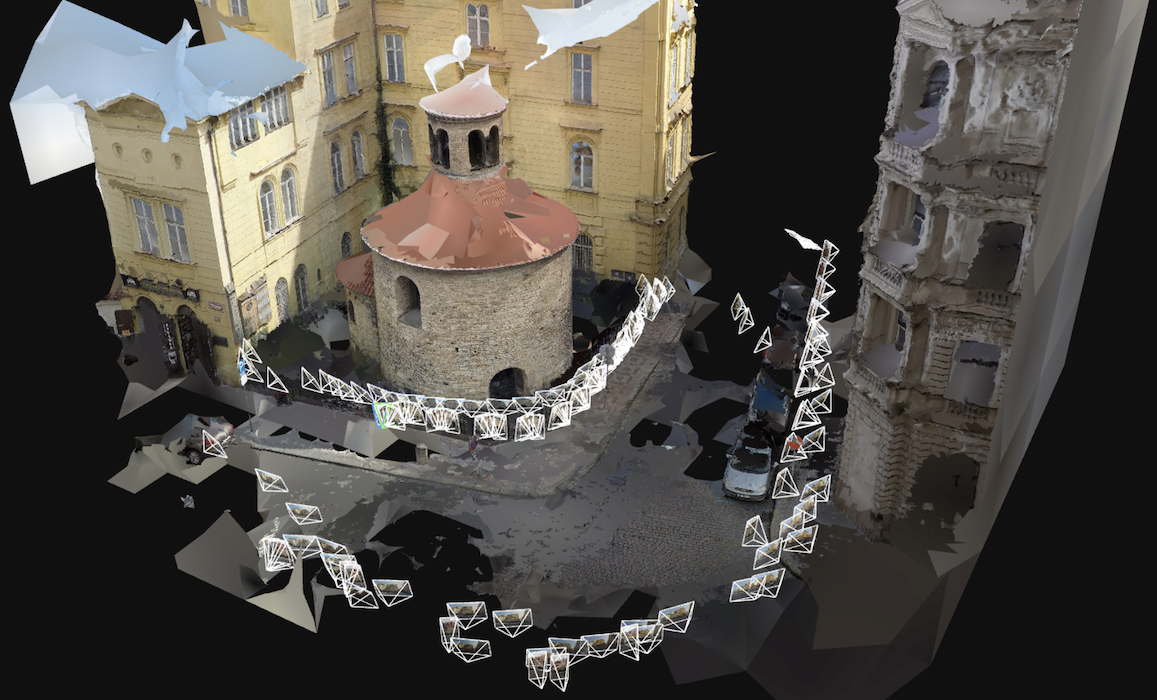}%
    \caption{Visualization of the \ROTUNDA scene. We show a textured mesh of the scene to provide a clearer visualization. We also show the poses of the 157 images of the dataset.}
    \label{fig:datasets_rotunda_3D}
\end{figure}

\newnewtext{We use 
\newtext{%
GeoCalib~\cite{veicht2024geocalib}, %
\newnewtext{a recent, state-of-the-art} end-to-end deep learning approach that} 
predicts camera intrinsics (focal length and a radial undistortion parameter) and the gravity direction from a single image or multiple images. 
\newtext{
GeoCalib first employs a convolutional neural network to infer visual cues \newnewtext{in the form of} a Perspective Field~\cite{jin2023perspective}, \newnewtext{storing} %
per-pixel up-vector and latitude \newnewtext{estimates and their uncertainties}. %
\newnewtext{The} camera parameters \newnewtext{that model the observations stored in} %
this Perspective Field are \newnewtext{then} found using \newnewtext{differentiable Levenberg-Marquart (LM)} %
optimization. If multiple images produced by a single camera are available, the shared intrinsics can be estimated jointly from the Perspective Fields of all images, resulting in better accuracy.}}

\newtext{GeoCalib simultaneously predicts the camera's focal length, a radial undistortion parameter \ZK{for one-parameter division model}, and the gravity direction. These camera parameter predictions can be used within a RANSAC pipeline as prior information, similarly to how the sampled undistortion parameters are utilized. 
We \newnewtext{ use these priors in three ways:} %
(1) we only use %
the predicted radial undistortion parameter, \newnewtext{instead of a sampled parameter}, %
and run the standard 7pt \F \ or 6pt \Ef \ solvers on image points undistorted by the predicted undistortion parameter. 
(2) we use %
both the predicted radial undistortion parameter and the focal length as priors. 
We use the focal length to calibrate the image points, and the undistorion parameter to undistort them. 
We then %
run the 5pt solver for calibrated pinhole cameras. %
(3) we use all predicted parameters to %
run the 3pt solver~\cite{ding2023revisiting, ding2020efficient} that estimates the relative pose between two cameras with known gravity directions.} 

\begin{figure}[t!]
    \centering
    \begin{tabular}{c@{\hskip 1mm}c}
    \includegraphics[width=30mm]{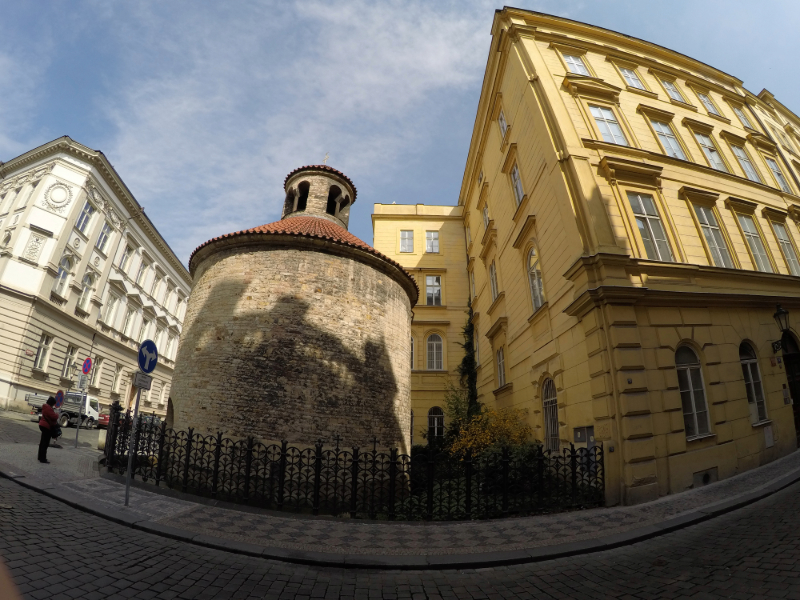} &
    \includegraphics[width=30mm]{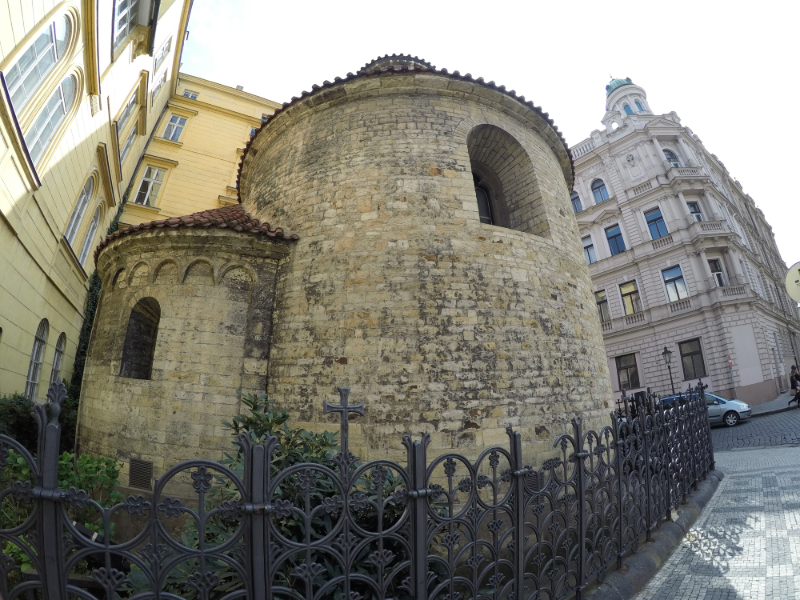} \\
    \includegraphics[width=30mm]{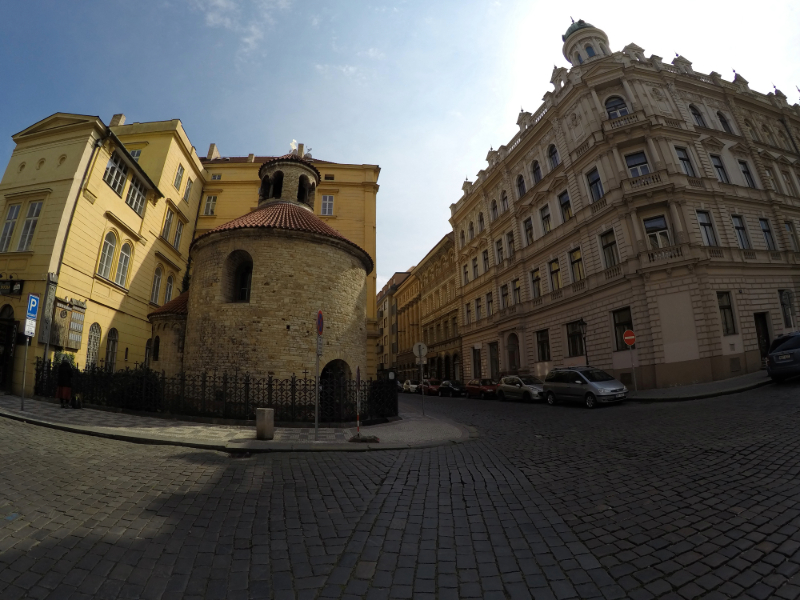} &
    \includegraphics[width=30mm]{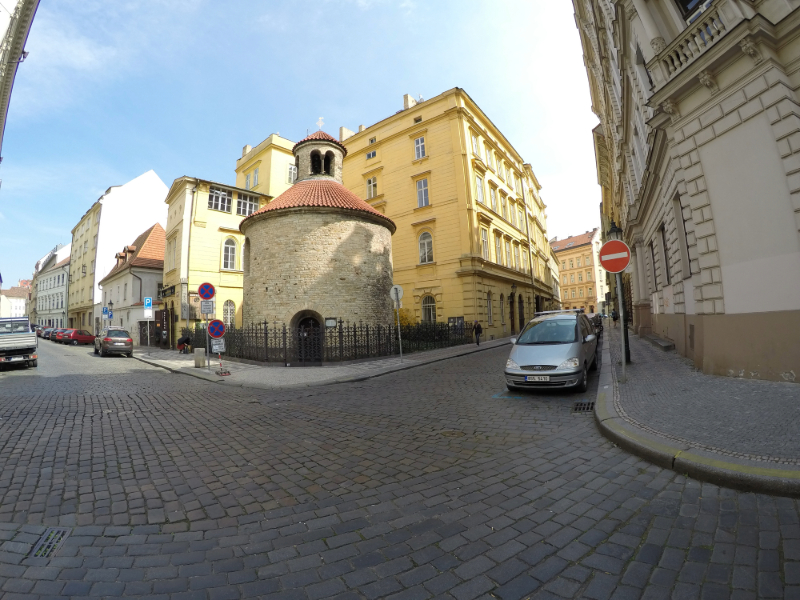}\\ 
    \end{tabular}
    \caption{Example images from the \ROTUNDA scene.}
    \label{fig:datasets2}
\end{figure}

\newnewtext{Compared to our sampling strategy, using learning-based priors has the potential to speed up the estimation process: 
\ZK{%
(1) The inferred undistortion parameter can be closer to the ground truth parameter than the sampled one(s). (2) We only need to test a single  parameter compared to several undistortion parameters that usually have to be evaluated by sampling-based strategy. (3) The inferred focal lengths and gravity directions simplify the relative pose problems that need to be solved.} 
However, a modern GPU is needed for GeoCalib, whereas the sampling strategy only requires CPU-based computations. 
Thus, the approach that uses learning-based priors might not always be applicable, \eg, it requires too much resources for robotics and on-device augmented reality applications, where energy consumption and battery capacity are limiting factors. 
Furthermore, the predictions by GeoCalib might not be accurate, especially if the input image(s) is very different from GeoCalib's training data. 
In such cases, our sampling-based strategy will still perform well.}

\section{Experiments}

\subsection{Datasets} 
We evaluate the different approaches for radial distortion relative pose estimation on four datasets: 
\ETH \cite{Schops_2017_CVPR}, %
\newtext{\PP~\cite{IMC2020}, \Euroc~\cite{Burri25012016},}
and our new benchmark, %
each covering different scenarios. 

The 
\ETH 
dataset 
was designed to evaluate (multi-view) stereo algorithms~\cite{Schops_2017_CVPR}. 
It covers indoor and outdoor scenes captured with a DSLR
camera. %
Ground truth poses were obtained by aligning the images to high-precision laser scans. 
\ETH provides undistorted images together with their intrinsic calibration. %
We use 2,037 image pairs from 12 \ETH scenes. %

\newtext{The \PP dataset~\cite{IMC2020} contains images extracted from iPhone 11 video sequences, in which both standard and wide-angle lenses are used. 
The authors provide ground truth poses obtained using
RealityCapture SfM software~\cite{RealityCapture}. The dataset features vegetation-rich scenes such as trees, ponds, sculptures, with different levels of zoom, and no people. We use 453 pairs of undistorted images from 3 \PP scenes. }
\ZK{We use undistorted images from \ETH and \PP datasets in experiments, where we synthetically distort them to simulate different scenarios, \eg scenario with images with different distortions downloaded from the Internet.}

\newtext{The \Euroc dataset~\cite{Burri25012016} is a widely used benchmark for visual-inertial odometry and SLAM, captured using a Micro Aerial Vehicle equipped with synchronized stereo cameras and an IMU. It features sequences recorded in indoor environments such as machine halls and office spaces, with varying levels of motion dynamics and lighting conditions. The dataset provides accurate ground truth from a motion capture system, making it ideal for evaluating pose estimation algorithms. We use 13,315 image pairs from 6 \Euroc scenes. The provided ground truth parameters are based on a radial-tangential distortion model rather than the division model, therefore, distortion error is not reported for this dataset. %
}

\begin{figure}[t!]
    \centering
    \includegraphics[width=0.95\columnwidth]{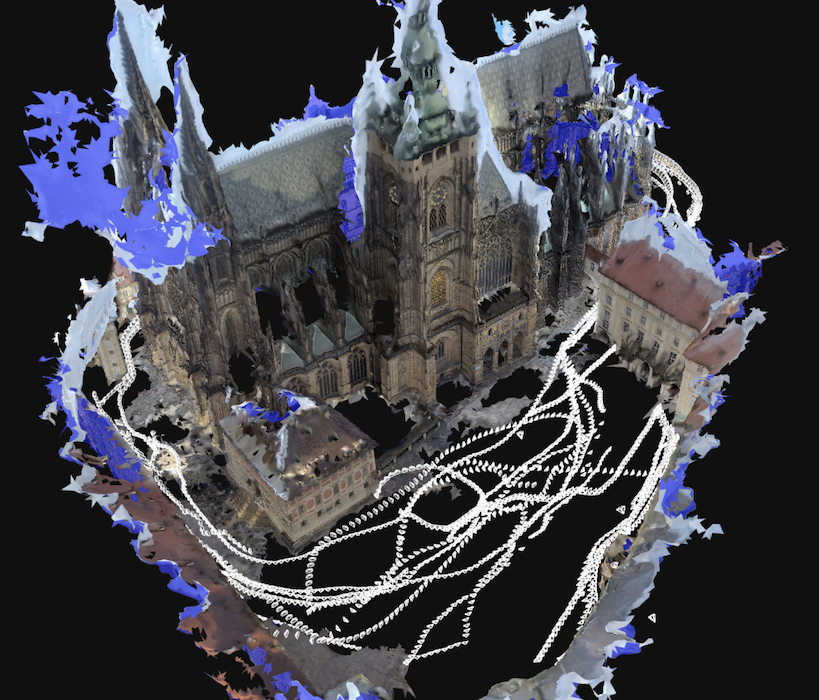}%
    \caption{Visualization of the \VITUS scene. We show a colored mesh of the scene to provide a clearer visualization. We also show the poses of the 2,734 images of the dataset.}
    \label{fig:datasets_cathedral_3D}
\end{figure}

\subsection{New Benchmark}

Existing datasets~\cite{Burri25012016, Schops_2017_CVPR} containing radially distorted images %
mostly involve only one or two different camera lenses with little %
variation in the undistortion parameters. 
Testing the sampling-based strategy on such images could be biased, as it would have been as good as the distance of the used sampled value from the one/two ground truth values. 
We thus created a new benchmark with a higher variation in the undistortion parameters,  
consisting of two scenes:  \ROTUNDA and {\fontfamily{cmtt}\selectfont CATHERAL}. %
For both scenes, we build upon previously recorded images~\cite{kukelova2015efficient,Sattler_2019_CVPR}, taken by GoPro cameras and kindly provided by the authors. 
For both scenes, we recorded additional images with different cameras \ZK{and, in addition, we downloaded some images from Flickr that depict \CATHEDRAL scene. }
To obtain ground truth poses and camera intrinsics including radial distortion for both the original and the newly added images, we used the RealityCapture software~\cite{RealityCapture}.
We configured RealityCapture to estimate the undistortion parameters for the images using the one-parameter division model that we use in all of our experiments, \ie, RealityCapture directly provides ground truth estimates for the undistortion parameters. 
We enforced that images taken by the same camera (with the same field of view) share the same intrinsic camera parameters. 
In the following, we briefly describe both scenes. 

\begin{figure}[t!]
    \centering
    \begin{tabular}{c@{\hskip 1mm}c}
    \includegraphics[width=27mm]{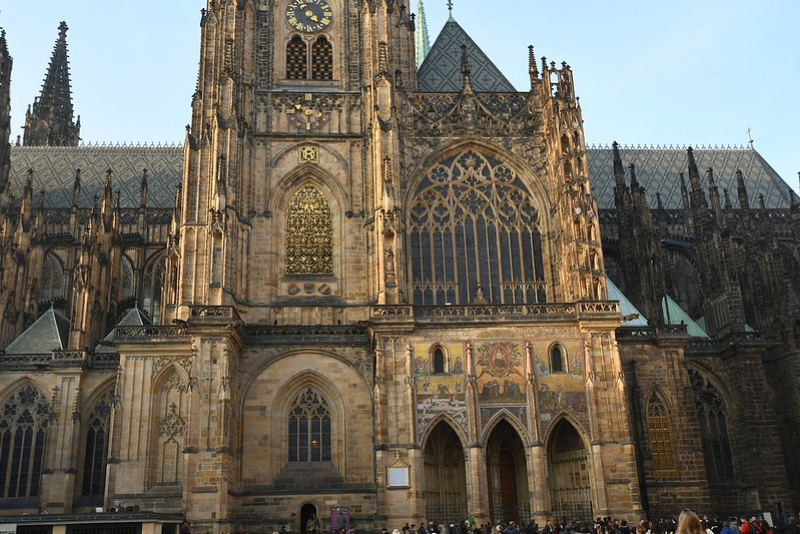} &
    \includegraphics[width=30mm]{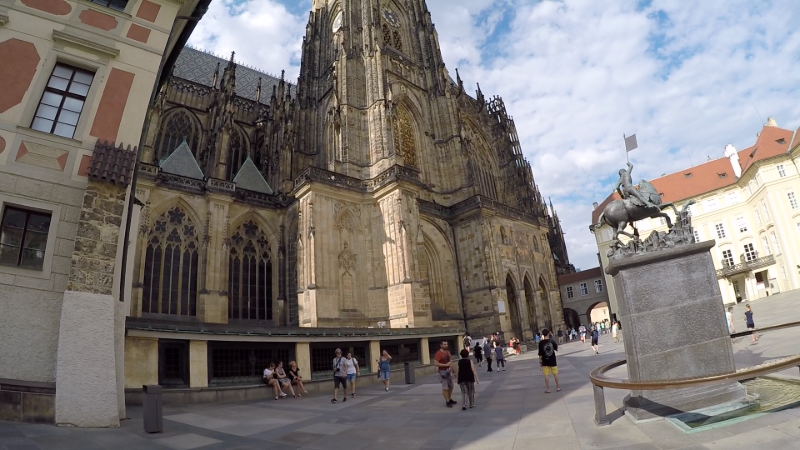}\\
    \includegraphics[width=27mm]{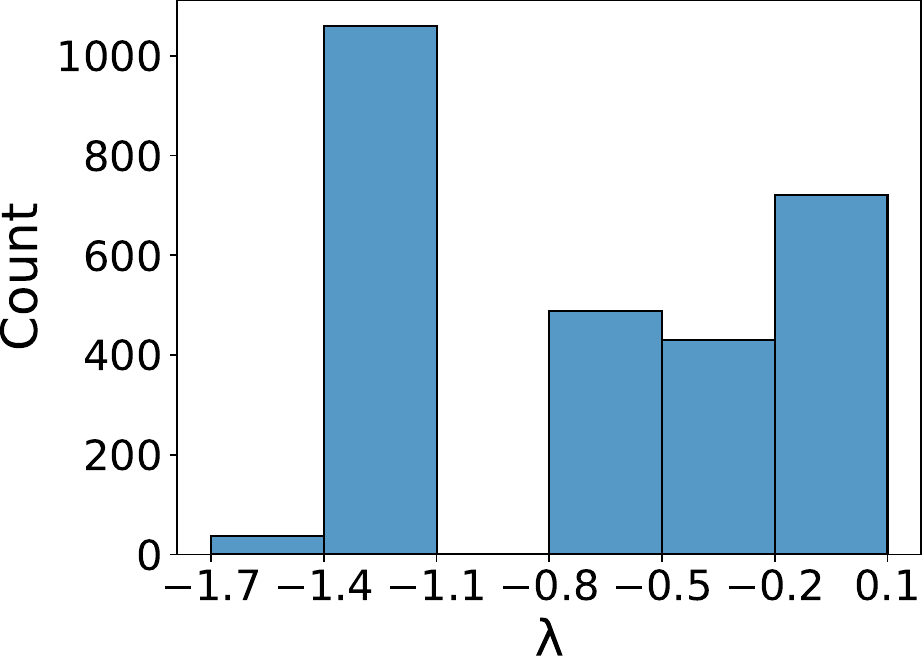} &
    \includegraphics[width=27mm]{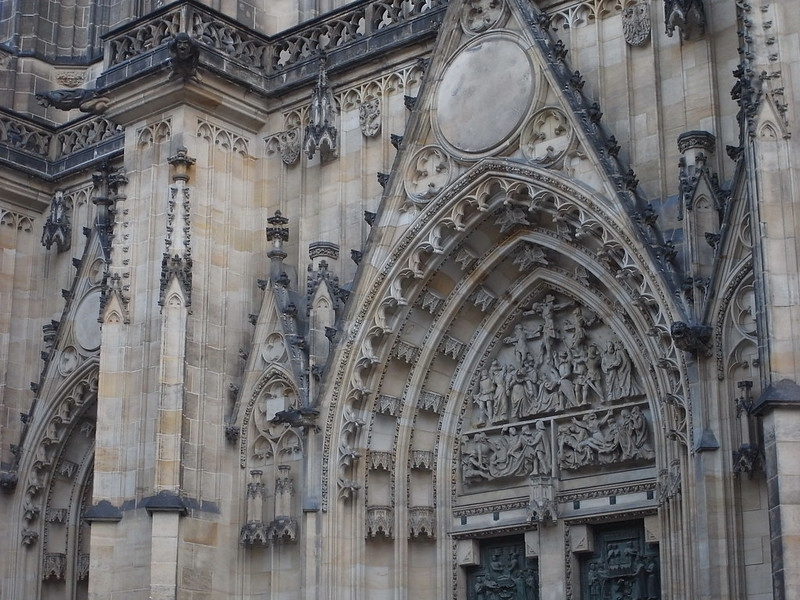} \\
    \end{tabular}
    \caption{Example images from the \VITUS scene. Distribution of $\lambda$ radial distortion parameters for the  \VITUS scene (a). The parameters were obtained by normalizing the ground truth parameters estimated by RealityCapture.}
    \label{fig:datasets3}
\end{figure}

The \ROTUNDA scene contains 157 outdoor images of a historical building captured by two mobile phone cameras (95 new images in total) and one GoPro camera (62 images, provided by the authors of~\cite{kukelova2015efficient}). 
The GoPro images were captured using two different settings that affect the field-of-view and image distortion. 
Overall, images were taken with four different $\lambda$ values: $\left\{-1.50, -0.81,  0.02,  0.09\right\}$ (ground truth values provided by RealityCapture after normalization). 
Fig.~\ref{fig:datasets_rotunda_3D} visualizes the \ROTUNDA scene by showing a textured mesh of the scene together with the camera poses of the recorded images. 
Figure~\ref{fig:datasets2} shows example images from the \ROTUNDA scene. 

The \VITUS scene contains 2,734 outdoor images of a historical cathedral, captured by two mobile phone cameras (708 new images in total), one GoPro camera (655 images, provided by the authors of~\cite{Sattler_2019_CVPR}), and an Insta360 Ace Pro camera (1,358 new images). Most of the images were extracted from videos captured while walking around the building. In addition, we are using 13 images from Flickr. %
The dataset contains images from cameras with 42 different $\lambda$ parameters. Their distribution is shown in Fig.~\ref{fig:datasets3} along with example images for
the \VITUS scene. %
Fig.~\ref{fig:datasets_cathedral_3D} visualizes the \VITUS scene by showing a colored mesh of the scene together with the camera poses of the recorded images. 

For our experiments, we use 3,424 image pairs with two different cameras (denoted as $\lambda_1 \neq \lambda_2$) and 1,795 image pairs captured by the same camera and thus with the shared intrinsics (denoted as $\lambda_1 = \lambda_2$) for \ROTUNDA scene and 10,000 %
sampled image pairs for both $\lambda_1 \neq \lambda_2$ and $\lambda_1 = \lambda_2$ for {\fontfamily{cmtt}\selectfont CATHERAL} scene. %

\subsection{Evaluation measures} Following~\cite{IMC2020}, given the ground truth and the estimated relative pose, we measure the rotation error and the translation error. 
The rotation error is defined as the  angle of the rotation matrix aligning the estimated with the ground truth rotation matrix. 
The translation error is defined as the angle between the estimated and the ground truth translation vector. 
Finally, the pose error is defined as the maximum of the rotation error and the translation error. We also measure the distortion error $\epsilon(\lambda)$ as the absolute distance between ground truth and estimated undistortion parameters \newtext{and the focal length error as $\xi(f) = \frac{|f_{gt} - f|}{f_{gt}}$, where $f$ is the estimated focal length and $f_{gt}$ is the ground truth. For the problem with two different cameras, we measure the distortion error as  $0.5 \cdot (\epsilon(\lambda_1)+\epsilon(\lambda_2))$ and the focal length error as $0.5 \cdot (\xi(f_1) + \xi(f_2))$.} 
We report the average (AVG) and median (MED) pose errors, as well as the area under the recall curve (AUC)
\newtext{at a $10^\circ$ threshold} on the pose error. 

\subsection{Experimental setup} 
\newtext{We obtained the correspondences between the images by matching SuperPoint~\cite{detone2018superpoint} features extracted on the images without resizing. We only kept at most 2048 features and matched them with LightGlue (LG)~\cite{lindenberger2023lightglue}. We only considered images with sufficient overlap quantified by the co-visibility constraint proposed in~\cite{IMC2020}. We retained only those image pairs that yielded a minimum of 20 matches.} \newtext{For \Euroc we kept at most 4096 features to ensure a sufficient number of matches as this indoor dataset is relatively textureless and more challenging.}

We evaluate the 8pt \Fk~\cite{Oskarsson_2021_CVPR}, 9pt \Fkk~\cite{Oskarsson_2021_CVPR}, and the non-minmal 9pt \Fk~(\cf Sec.~\ref{sec:ref_solvers_equal}), and  10pt \Fkk{}~\cite{kukelova2015efficient} solvers (\cf Sec.~\ref{sec:rd_solvers}) in %
RANSAC, and  the sampling strategies that combine the 6pt \Ef~\cite{larsson2017efficient}, and the 7pt \F~\cite{hartley2003multiple} solver with a set of pre-defined undistortion parameters (\cf Sec.~\ref{sec:rd_solvers:sampling}). We denote the latter by appending the list of parameters, \eg, $\{0, -0.6, -1.2\}$, to the solver configuration. \newtext{Additionally, we evaluate the learning-based prior strategy which combines the 7pt \F~\cite{hartley2003multiple}, 6pt \Ef~\cite{larsson2017efficient}, 5pt \E~\cite{nister2004efficient}, and 3pt \E~\cite{ding2023revisiting, ding2020efficient} solvers with the camera parameters predicted using GeoCalib (\cf Sec.~\ref{sec:learning-based_strategy}). The GeoCalib predictions are obtained by running the network with its default inference settings, followed by 30 iterations (which is the default number of iterations) of LM optimization to refine the estimated camera intrinsics and gravity direction. For the case when the two cameras share intrinsics ($\lambda_1\,=\,\lambda_2$), we utilize GeoCalib's multi-image optimization setting, which jointly optimizes the shared focal length and radial undistortion parameter of a pair of images while independently refining the gravity direction of each image. In contrast, for the case of two different cameras ($\lambda_1\,\not=\,\lambda_2$), each image is processed independently, and camera parameters are predicted using the default single-image inference pipeline of GeoCalib. GeoCalib inference with the multi-image optimization setting for pairs of images and 30 iterations runs on average $\sim 380 ms$, while for a single-image, inference runs on average $\sim 185ms$ \vk{on an NVIDIA A100 (a high-end server grade GPU)}.} \ZK{We also add an ablation study that analyzes the impact of the number of iterations of LM optimization on run-time and pose estimation accuracy.} %

We integrate the solvers and strategies %
into PoseLib~\cite{PoseLib}.
The LO step in PoseLib relies on Levenberg-Marquardt (LM) optimization of the truncated Tangent Sampson Error~\cite{Terekhov_2023_ICCV}, starting from the estimate provided by the minimal solver, sampling strategy, \newtext{or the learning-based prior.}
The pose and intrinsics returned by solvers are further polished %
by LM optimization over all inliers. \newtext{For the learning-based prior strategy we evaluate two different LO settings. We either optimized all camera parameters, or leave the parameters estimated by GeoCalib fixed. For each method we denote which parameters were refined in LO.} \newtext{For solvers which produce the fundamental matrix we decompose it using the Bougnoux formula for two different cameras~\cite{bougnoux1998projective} and the Sturm's formula for cameras with shared intrinsics~\cite{sturm2001focal} into the pose and the focal lengths. We use the closed-form formulas due to their speed since fundamental matrices need to be decomposed for each provided solution. For both decomposition and local optimization we assume that the principal point is fixed in the image center.}

To determine which points are inliers we use the Tangent Sampson Error~\cite{Terekhov_2023_ICCV} with a fixed 3px threshold. 
Using the Tangent Sampson Error is important since the standard Sampson Error in undistorted images leads to a radial bias in the optimization~\cite{Terekhov_2023_ICCV}.

We use normalized image coordinates from the range %
$[-0.5, 0.5]^2$, 
obtained by subtracting the image center
and dividing by the length of the longer image side. 
For this normalization, the undistortion parameter should be greater than $-2$, as otherwise the distortion would mirror the image. 
In RANSAC, we discard models with radial distortion 
outside of the plausible range $\left[-2.0, 0.5\right]$.

\begin{table*}[ht]
    \centering
    \caption{\textbf{Prior knowledge about cameras}: results on all scenes of the \ETH dataset, using Poselib RANSAC for synthetic scenario A - \textit{Wild} (\cf Sec.~\ref{sec:scenarioA}).
    The table shows the average and median pose error in degrees; the Area Under Recall Curve (AUC) at 10$^\circ$; the average and median absolute error $\epsilon(\lambda)$ of the undistortion parameter; \newtext{the average and median focal length error $\xi(f)$} and the average runtime of RANSAC. We highlight the \bestword{best} and \secondword{second-best} results. 
    }
    \resizebox{\textwidth}{!}{\begin{tabular}{ c | r c c | c c c | c c | c c | c}
    \toprule
    & & & & \multicolumn{7}{c}{Poselib - \ETH - Synth A} \\
    \midrule
    & Minimal & Refinement & Sample & AVG $(^\circ)$ $\downarrow$ & MED $(^\circ)$ $\downarrow$ & AUC@10 $\uparrow$ & AVG $\epsilon(\lambda)$ $\downarrow$ & MED $\epsilon(\lambda)$ $\downarrow$  & AVG $\xi(f)$ $\downarrow$ & MED $\xi(f)$ $\downarrow$ & Time (ms) $\downarrow$ \\
    \midrule
    \multirow{25}{*}{\rotatebox[origin=c]{90}{$\lambda_1 = \lambda_2$}}     & 9pt \Fkk & $\R,\tvec,f_1,f_2,\lambda_1,\lambda_2$ & \ding{55} & 42.46 & \phantom{1}7.42 & 0.40 & 0.47 & 0.13 & 0.36 & 0.21 & 826.20 \\
    & 10pt \Fkk & $\R,\tvec,f_1,f_2,\lambda_1,\lambda_2$ & \ding{55} & 38.45 & \phantom{1}5.87 & 0.44 & 0.41 & 0.11 & 0.29 & 0.19 & 129.70 \\
    & 8pt \Fk & $\R,\tvec,f,\lambda$ & \ding{55} & 31.78 & \phantom{1}2.87 & 0.56 & 0.38 & 0.06 & 0.39 & 0.09 & 579.29 \\
    & 9pt \Fk & $\R,\tvec,f,\lambda$ & \ding{55} & 31.35 & \phantom{1}2.97 & 0.56 & 0.43 & 0.07 & 0.36 & 0.10 & 219.00 \\
    & 7pt \F & $\R,\tvec,f_1,f_2$ & $\lambda$ = 0 & 50.51 & 23.74 & 0.15 & 0.86 & 0.87 & 0.84 & 0.54 & \phantom{1}91.89 \\
    & 7pt \F & $\R,\tvec,f_1,f_2,\lambda_1,\lambda_2$ & $\lambda = 0$ & 43.13 & \phantom{1}7.41 & 0.40 & 0.40 & 0.12 & 0.35 & 0.23 & \phantom{1}89.33 \\
    & 7pt \F & $\R,\tvec,f_1,f_2,\lambda_1,\lambda_2$ & $\lambda \in \{0.0, -0.6, -1.2\}$ & 35.36 & \phantom{1}5.66 & 0.45 & 0.29 & 0.11 & 0.33 & 0.19 & 136.46 \\
    & 6pt \Ef & $\R,\tvec,f,\lambda$ & $\lambda = 0$ & 15.13 & \phantom{1}2.05 & 0.65 & 0.21 & 0.05 & 0.48 & 0.06 & 104.25 \\
    & 6pt \Ef & $\R,\tvec,f,\lambda$ & $\lambda \in \{0.0, -0.6, -1.2\}$ & \cellcolor{best}{13.86} & \phantom{1}\cellcolor{best}{1.76} & \cellcolor{best}{0.68} & 0.15 & \cellcolor{second}{0.05} & 0.50 & \cellcolor{best}{0.05} & 142.94 \\
    & 7pt \F & $\R,\tvec,f_1,f_2$ & GeoCalib - $\lambda$ & 36.23 & \phantom{1}8.74 & 0.35 & 0.24 & 0.13 & 0.41 & 0.28 & \phantom{1}81.56 \\
    & 7pt \F & $\R,\tvec,f_1,f_2,\lambda_1,\lambda_2$ &  GeoCalib - $\lambda$ & 35.60 & \phantom{1}5.95 & 0.43 & 0.29 & 0.10 & 0.33 & 0.19 & \phantom{1}90.80 \\
    & 6pt \Ef & $\R,\tvec,f$ & GeoCalib - $\lambda$ & 15.63 & \phantom{1}2.49 & 0.63 & \cellcolor{second}{0.14} & 0.06 & 0.66 & 0.07 & \phantom{1}96.06 \\
    & 6pt \Ef & $\R,\tvec,f,\lambda$ & GeoCalib - $\lambda$ & \cellcolor{second}{14.45} & \phantom{1}\cellcolor{second}{1.96} & \cellcolor{second}{0.67} & \cellcolor{best}{0.13} & \cellcolor{best}{0.04} & 0.54 & \cellcolor{second}{0.05} & 102.33 \\
    & 5pt \E & $\R, \tvec$ & GeoCalib - $\lambda,f$ & 19.01 & \phantom{1}3.87 & 0.54 & 0.20 & 0.10 & \cellcolor{best}{0.19} & 0.13 & \phantom{1}\cellcolor{second}{42.68} \\
    & 5pt \E & $\R,\tvec,f,\lambda$ & GeoCalib - $\lambda,f$ & 18.70 & \phantom{1}3.52 & 0.56 & 0.23 & 0.09 & 0.22 & 0.13 & 114.52 \\
    & 3pt \E & $\R,\tvec$ & GeoCalib - $\lambda,f,\g$ & 22.12 & \phantom{1}4.09 & 0.52 & 0.20 & 0.10 & \cellcolor{second}{0.19} & 0.13 & \phantom{1}\cellcolor{best}{40.93} \\
    & 3pt \E & $\R,\tvec,f,\lambda$ & GeoCalib - $\lambda,f,\g$ & 25.00 & \phantom{1}4.32 & 0.51 & 0.26 & 0.10 & 0.24 & 0.15 & 120.30 \\
    \midrule
    \midrule
    \multirow{16}{*}{\rotatebox[origin=c]{90}{$\lambda_1 \neq \lambda_2$}}     & 9pt \Fkk & $\R,\tvec,f_1,f_2,\lambda_1,\lambda_2$ & \ding{55} & 46.27 & \phantom{1}8.80 & 0.38 & 0.47 & 0.13 & 0.39 & 0.24 & 743.74 \\
    & 10pt \Fkk & $\R,\tvec,f_1,f_2,\lambda_1, \lambda_2$ & \ding{55} & 37.75 & \phantom{1}6.07 & 0.43 & 0.40 & 0.10 & 0.28 & 0.18 & 132.32 \\
    & 7pt \F & $\R,\tvec,f_1, f_2$ & $\lambda_1 = \lambda_2$ = 0 & 54.77 & 29.09 & 0.10 & 0.88 & 0.87 & 0.81 & 0.60 & 101.93 \\
    & 7pt \F & $\R,\tvec,f_1, f_2, \lambda_1, \lambda_2$ & $\lambda_1 = \lambda_2 = 0  $ & 46.21 & \phantom{1}8.79 & 0.37 & 0.40 & 0.12 & 0.35 & 0.24 & \phantom{1}94.50 \\
     & 7pt \F & $\R,\tvec,f_1, f_2, \lambda_1, \lambda_2$ & $\lambda_1, \lambda_2 \in \{0.0, -0.6, -1.2\} $ & \cellcolor{second}{32.97} & \phantom{1}6.30 & 0.42 & 0.27 & 0.10 & 0.33 & 0.20 & 138.85 \\
    & 7pt \F & $\R,\tvec,f_1, f_2$ & GeoCalib - $\lambda_1, \lambda_2$ & 37.74 & 10.60 & 0.30 & 0.30 & 0.19 & 0.43 & 0.31 & \phantom{1}83.60 \\
    & 7pt \F & $\R,\tvec,f_1,f_2,\lambda_1, \lambda_2$ &  GeoCalib - $\lambda_1, \lambda_2$ & 35.89 & \phantom{1}6.62 & 0.42 & 0.27 & \cellcolor{second}{0.10} & 0.33 & 0.20 & \phantom{1}95.03 \\
    & 5pt \E & $\R, \tvec$ & GeoCalib - $\lambda_1, \lambda_2,f_1, f_2$ & 33.15 & \phantom{1}8.96 & 0.33 & \cellcolor{second}{0.23} & 0.16 & \cellcolor{second}{0.24} & 0.19 & \phantom{1}\cellcolor{second}{55.62} \\
    & 5pt \E & $\R,\tvec,f_1, f_2,\lambda_1, \lambda_2$ & GeoCalib - $\lambda_1,\lambda_2,f_1,f_2$ & \cellcolor{best}{29.44} & \phantom{1}\cellcolor{best}{4.75} & \cellcolor{best}{0.49} & \cellcolor{best}{0.23} & \cellcolor{best}{0.09} & \cellcolor{best}{0.24} & \cellcolor{best}{0.16} & 123.20 \\
    & 3pt \E & $\R,\tvec$ & GeoCalib - $\lambda_1, \lambda_2,f_1, f_2,\g_1,\g_2$ & 34.98 & \phantom{1}9.18 & 0.32 & 0.23 & 0.16 & 0.24 & 0.19 & \phantom{1}\cellcolor{best}{43.36} \\
    & 3pt \E & $\R,\tvec,f_1,f_2, \lambda_1, \lambda_2 $ & GeoCalib - $\lambda_1, \lambda_2, f_1, f_2, \g_1, \g_2$ & 35.16 & \phantom{1}\cellcolor{second}{5.18} & \cellcolor{second}{0.47} & 0.24 & 0.10 & 0.24 & \cellcolor{second}{0.16} & 123.34 \\
    \bottomrule
\end{tabular}
}    
    \label{tab:poselib_eth3d_synth_A}
\end{table*}

\begin{table*}[ht]
    \centering
    \caption{\textbf{Prior knowledge about cameras}: results on all scenes of the \ETH dataset, using Poselib RANSAC for synthetic scenario B - \textit{Small Distortion} (\cf Sec.~\ref{sec:scenarioB}). 
    The reported statistics are the same as in Tab.~\ref{tab:poselib_eth3d_synth_A}.}
    \resizebox{\textwidth}{!}{\begin{tabular}{ c | r c c | c c c | c c | c c | c}
    \toprule
    & & & & \multicolumn{7}{c}{Poselib - \ETH - Synth B} \\
    \midrule
    & Minimal & Refinement & Sample & AVG $(^\circ)$ $\downarrow$ & MED $(^\circ)$ $\downarrow$ & AUC@10 $\uparrow$ & AVG $\epsilon(\lambda)$ $\downarrow$ & MED $\epsilon(\lambda)$ $\downarrow$  & AVG $\xi(f)$ $\downarrow$ & MED $\xi(f)$ $\downarrow$ & Time (ms) $\downarrow$ \\
    \midrule
    \multirow{21}{*}{\rotatebox[origin=c]{90}{$\lambda_1 = \lambda_2$}}     & 9pt \Fkk & $\R,\tvec,f_1,f_2,\lambda_1,\lambda_2$ & \ding{55} & 34.15 & 5.60 & 0.45 & 0.27 & 0.07 & 0.35 & 0.16 & 824.68 \\
    & 10pt \Fkk & $\R,\tvec,f_1,f_2,\lambda_1,\lambda_2$ & \ding{55} & 35.87 & 5.58 & 0.46 & 0.31 & 0.08 & 0.34 & 0.17 & 127.75 \\
    & 8pt \Fk & $\R,\tvec,f,\lambda$ & \ding{55} & 29.49 & 2.24 & 0.59 & 0.29 & 0.04 & 0.46 & 0.06 & 650.20 \\
    & 9pt \Fk & $\R,\tvec,f,\lambda$ & \ding{55} & 29.47 & 2.28 & 0.60 & 0.32 & 0.04 & 0.39 & 0.07 & 228.97 \\
    & 7pt \F & $\R,\tvec,f_1,f_2$ & $\lambda$ = 0 & 33.92 & 7.43 & 0.38 & 0.15 & 0.15 & 0.52 & 0.23 & \phantom{1}80.98 \\
    & 7pt \F & $\R,\tvec,f_1,f_2,\lambda_1,\lambda_2$ & $\lambda = 0$ & 32.36 & 4.90 & 0.47 & 0.21 & 0.07 & 0.32 & 0.17 & \phantom{1}83.15 \\
    & 6pt \Ef & $\R,\tvec,f,\lambda$ & $\lambda = 0$ & \cellcolor{best}{12.26} & \cellcolor{best}{1.46} & \cellcolor{second}{0.72} & 0.09 & \cellcolor{second}{0.03} & 0.43 & \cellcolor{best}{0.04} & \phantom{1}95.47 \\
    & 7pt \F & $\R,\tvec,f_1,f_2$ & GeoCalib - $\lambda$ & 31.37 & 5.11 & 0.47 & 0.08 & 0.05 & 0.39 & 0.19 & \phantom{1}76.22 \\
    & 7pt \F & $\R,\tvec,f_1,f_2,\lambda_1,\lambda_2$ &  GeoCalib - $\lambda$ & 31.28 & 4.90 & 0.47 & 0.19 & 0.06 & 0.34 & 0.16 & \phantom{1}82.40 \\
    & 6pt \Ef & $\R,\tvec,f$ & GeoCalib - $\lambda$ & \cellcolor{second}{12.67} & 1.58 & 0.71 & 0.06 & 0.04 & 0.63 & 0.04 & \phantom{1}90.57 \\
    & 6pt \Ef & $\R,\tvec,f,\lambda$ & GeoCalib - $\lambda$ & 13.04 & \cellcolor{second}{1.47} & \cellcolor{best}{0.72} & 0.09 & \cellcolor{best}{0.03} & 0.59 & \cellcolor{second}{0.04} & \phantom{1}94.37 \\
    & 5pt \E & $\R, \tvec$ & GeoCalib - $\lambda,f$ & 23.11 & 3.93 & 0.53 & \cellcolor{second}{0.05} & 0.04 & \cellcolor{second}{0.25} & 0.17 & \phantom{1}\cellcolor{second}{39.55} \\
    & 5pt \E & $\R,\tvec,f,\lambda$ & GeoCalib - $\lambda,f$ & 22.53 & 3.81 & 0.54 & 0.17 & 0.07 & 0.28 & 0.14 & 108.85 \\
    & 3pt \E & $\R,\tvec$ & GeoCalib - $\lambda,f,\g$ & 27.72 & 4.32 & 0.50 & \cellcolor{best}{0.05} & 0.04 & \cellcolor{best}{0.25} & 0.17 & \phantom{1}\cellcolor{best}{37.48} \\
    & 3pt \E & $\R,\tvec,f,\lambda$ & GeoCalib - $\lambda,f,\g$ & 28.56 & 4.29 & 0.51 & 0.21 & 0.07 & 0.31 & 0.15 & 115.08 \\
    \midrule
    \midrule
    \multirow{14}{*}{\rotatebox[origin=c]{90}{$\lambda_1 \neq \lambda_2$}}     & 9pt \Fkk & $\R,\tvec,f_1,f_2,\lambda_1,\lambda_2$ & \ding{55} & 36.79 & 5.17 & 0.47 & 0.30 & 0.07 & 0.31 & 0.16 & 747.69 \\
    & 10pt \Fkk & $\R,\tvec,f_1,f_2,\lambda_1, \lambda_2$ & \ding{55} & 34.73 & 4.95 & 0.48 & 0.32 & 0.07 & \cellcolor{best}{0.30} & \cellcolor{best}{0.15} & 130.13 \\
    & 7pt \F & $\R,\tvec,f_1, f_2$ & $\lambda_1 = \lambda_2$ = 0 & 34.63 & 9.14 & 0.34 & 0.15 & 0.15 & 0.52 & 0.26 & \phantom{1}83.45 \\
    & 7pt \F & $\R,\tvec,f_1, f_2, \lambda_1, \lambda_2$ & $\lambda_1 = \lambda_2 = 0  $ & 32.01 & 4.86 & 0.48 & 0.20 & 0.07 & 0.31 & 0.16 & \phantom{1}85.37 \\
    & 7pt \F & $\R,\tvec,f_1, f_2$ & GeoCalib - $\lambda_1, \lambda_2$ & \cellcolor{second}{31.50} & 5.83 & 0.44 & 0.09 & 0.06 & 0.38 & 0.21 & \phantom{1}76.66 \\
    & 7pt \F & $\R,\tvec,f_1,f_2,\lambda_1, \lambda_2$ &  GeoCalib - $\lambda_1, \lambda_2$ & \cellcolor{best}{30.54} & \cellcolor{best}{4.51} & \cellcolor{best}{0.48} & 0.18 & 0.06 & 0.30 & 0.16 & \phantom{1}86.09 \\
    & 5pt \E & $\R, \tvec$ & GeoCalib - $\lambda_1, \lambda_2,f_1, f_2$ & 36.26 & 8.63 & 0.34 & \cellcolor{best}{0.07} & \cellcolor{best}{0.06} & 0.30 & 0.22 & \phantom{1}\cellcolor{second}{50.42} \\
    & 5pt \E & $\R,\tvec,f_1, f_2,\lambda_1, \lambda_2$ & GeoCalib - $\lambda_1,\lambda_2,f_1,f_2$ & 33.22 & \cellcolor{second}{4.72} & \cellcolor{second}{0.48} & 0.17 & 0.06 & \cellcolor{second}{0.30} & \cellcolor{second}{0.15} & 117.38 \\
    & 3pt \E & $\R,\tvec$ & GeoCalib - $\lambda_1, \lambda_2,f_1, f_2,\g_1,\g_2$ & 38.66 & 8.86 & 0.34 & \cellcolor{second}{0.07} & \cellcolor{second}{0.06} & 0.30 & 0.22 & \phantom{1}\cellcolor{best}{42.19} \\
    & 3pt \E & $\R,\tvec,f_1,f_2, \lambda_1, \lambda_2 $ & GeoCalib - $\lambda_1, \lambda_2, f_1, f_2, \g_1, \g_2$ & 37.82 & 5.21 & 0.46 & 0.19 & 0.07 & 0.30 & 0.17 & 119.90 \\
    \bottomrule
\end{tabular}
}    
    \label{tab:poselib_eth3d_synth_B}
\end{table*}

\begin{table*}[ht]
    \centering
    \caption{\textbf{Prior knowledge about cameras}: results on all scenes of the \ETH dataset, using Poselib RANSAC for synthetic scenario C - \textit{Visible distortion} (\cf Sec.~\ref{sec:scenarioC}). %
    The reported statistics are the same as in Tab.~\ref{tab:poselib_eth3d_synth_A}.}
    \resizebox{\textwidth}{!}{\begin{tabular}{ c | r c c | c c c | c c | c c | c}
    \toprule
    & & & & \multicolumn{7}{c}{Poselib - \ETH - Synth C} \\
    \midrule
    & Minimal & Refinement & Sample & AVG $(^\circ)$ $\downarrow$ & MED $(^\circ)$ $\downarrow$ & AUC@10 $\uparrow$ & AVG $\epsilon(\lambda)$ $\downarrow$ & MED $\epsilon(\lambda)$ $\downarrow$  & AVG $\xi(f)$ $\downarrow$ & MED $\xi(f)$ $\downarrow$ & Time (ms) $\downarrow$ \\
    \midrule
    \multirow{25}{*}{\rotatebox[origin=c]{90}{$\lambda_1 = \lambda_2$}}     & 9pt \Fkk & $\R,\tvec,f_1,f_2,\lambda_1,\lambda_2$ & \ding{55} & 45.09 & \phantom{1}8.93 & 0.37 & 0.52 & 0.14 & 0.38 & 0.24 & 758.63 \\
    & 10pt \Fkk & $\R,\tvec,f_1,f_2,\lambda_1,\lambda_2$ & \ding{55} & 37.67 & \phantom{1}6.14 & 0.43 & 0.45 & 0.12 & 0.29 & 0.19 & 129.42 \\
    & 8pt \Fk & $\R,\tvec,f,\lambda$ & \ding{55} & 31.36 & \phantom{1}2.97 & 0.56 & 0.38 & 0.07 & 0.37 & 0.10 & 559.78 \\
    & 9pt \Fk & $\R,\tvec,f,\lambda$ & \ding{55} & 32.45 & \phantom{1}3.03 & 0.55 & 0.43 & 0.07 & 0.36 & 0.11 & 213.17 \\
    & 7pt \F & $\R,\tvec,f_1,f_2$ & $\lambda$ = 0 & 39.03 & 10.00 & 0.32 & 0.29 & 0.17 & 0.42 & 0.31 & \phantom{1}82.38 \\
    & 7pt \F & $\R,\tvec,f_1,f_2,\lambda_1,\lambda_2$ & $\lambda = -0.9$ & 38.04 & \phantom{1}6.48 & 0.42 & 0.32 & 0.11 & 0.32 & 0.21 & \phantom{1}93.26 \\
    & 7pt \F & $\R,\tvec,f_1,f_2,\lambda_1,\lambda_2$ & $\lambda \in \{-0.6, -0.9, -1.2\}$ & 32.98 & \phantom{1}5.93 & 0.44 & 0.30 & 0.10 & 0.33 & 0.21 & 136.08 \\
    & 6pt \Ef & $\R,\tvec,f,\lambda$ & $\lambda = -0.9$ & \cellcolor{best}{13.04} & \phantom{1}\cellcolor{second}{1.83} & \cellcolor{second}{0.69} & 0.15 & 0.05 & 0.38 & 0.05 & 105.68 \\
    & 6pt \Ef & $\R,\tvec,f,\lambda$ & $\lambda \in \{-0.6, -0.9, -1.2\}$ & \cellcolor{second}{13.08} & \phantom{1}\cellcolor{best}{1.81} & \cellcolor{best}{0.69} & \cellcolor{second}{0.15} & \cellcolor{second}{0.05} & 0.44 & \cellcolor{best}{0.05} & 140.36 \\
    & 7pt \F & $\R,\tvec,f_1,f_2$ & GeoCalib - $\lambda$ & 36.87 & \phantom{1}9.57 & 0.32 & 0.30 & 0.17 & 0.44 & 0.31 & \phantom{1}81.31 \\
    & 7pt \F & $\R,\tvec,f_1,f_2,\lambda_1,\lambda_2$ &  GeoCalib - $\lambda$ & 36.67 & \phantom{1}6.89 & 0.41 & 0.31 & 0.11 & 0.36 & 0.21 & \phantom{1}92.08 \\
    & 6pt \Ef & $\R,\tvec,f$ & GeoCalib - $\lambda$ & 14.80 & \phantom{1}2.88 & 0.60 & 0.17 & 0.08 & 0.65 & 0.09 & \phantom{1}96.50 \\
    & 6pt \Ef & $\R,\tvec,f,\lambda$ & GeoCalib - $\lambda$ & 13.27 & \phantom{1}1.87 & 0.68 & \cellcolor{best}{0.14} & \cellcolor{best}{0.05} & 0.56 & \cellcolor{second}{0.05} & 103.18 \\
    & 5pt \E & $\R, \tvec$ & GeoCalib - $\lambda,f$ & 15.54 & \phantom{1}3.61 & 0.57 & 0.25 & 0.16 & \cellcolor{best}{0.17} & 0.12 & \phantom{1}\cellcolor{second}{42.40} \\
    & 5pt \E & $\R,\tvec,f,\lambda$ & GeoCalib - $\lambda,f$ & 16.10 & \phantom{1}3.41 & 0.57 & 0.25 & 0.09 & 0.20 & 0.14 & 114.67 \\
    & 3pt \E & $\R,\tvec$ & GeoCalib - $\lambda,f,\g$ & 19.84 & \phantom{1}3.72 & 0.55 & 0.25 & 0.16 & \cellcolor{second}{0.17} & 0.12 & \phantom{1}\cellcolor{best}{41.27} \\
    & 3pt \E & $\R,\tvec,f,\lambda$ & GeoCalib - $\lambda,f,\g$ & 25.00 & \phantom{1}3.92 & 0.53 & 0.27 & 0.11 & 0.22 & 0.14 & 120.69 \\
    \midrule
    \midrule
    \multirow{16}{*}{\rotatebox[origin=c]{90}{$\lambda_1 \neq \lambda_2$}}     & 9pt \Fkk & $\R,\tvec,f_1,f_2,\lambda_1,\lambda_2$ & \ding{55} & 45.39 & \phantom{1}8.91 & 0.38 & 0.49 & 0.13 & 0.35 & 0.23 & 732.84 \\
    & 10pt \Fkk & $\R,\tvec,f_1,f_2,\lambda_1, \lambda_2$ & \ding{55} & 40.36 & \phantom{1}6.47 & 0.42 & 0.44 & 0.11 & 0.28 & 0.19 & 130.61 \\
    & 7pt \F & $\R,\tvec,f_1, f_2$ & $\lambda_1 = \lambda_2$ = 0 & 41.09 & 12.78 & 0.25 & 0.38 & 0.26 & 0.50 & 0.35 & \phantom{1}85.74 \\
    & 7pt \F & $\R,\tvec,f_1, f_2, \lambda_1, \lambda_2$ & $\lambda_1 = \lambda_2 = -0.9  $ & 37.97 & \phantom{1}6.66 & 0.41 & 0.30 & 0.11 & 0.33 & 0.21 & \phantom{1}93.59 \\
     & 7pt \F & $\R,\tvec,f_1, f_2, \lambda_1, \lambda_2$ & $\lambda_1, \lambda_2 \in \{-0.6, -0.9, -1.2\} $ & 34.19 & \phantom{1}6.14 & 0.43 & 0.28 & \cellcolor{second}{0.10} & 0.32 & 0.21 & 136.95 \\
    & 7pt \F & $\R,\tvec,f_1, f_2$ & GeoCalib - $\lambda_1, \lambda_2$ & 39.49 & 12.26 & 0.26 & 0.37 & 0.25 & 0.45 & 0.35 & \phantom{1}84.80 \\
    & 7pt \F & $\R,\tvec,f_1,f_2,\lambda_1, \lambda_2$ &  GeoCalib - $\lambda_1, \lambda_2$ & 37.28 & \phantom{1}6.82 & 0.41 & 0.30 & 0.10 & 0.31 & 0.21 & \phantom{1}93.57 \\
    & 5pt \E & $\R, \tvec$ & GeoCalib - $\lambda_1, \lambda_2,f_1, f_2$ & \cellcolor{second}{26.74} & \phantom{1}7.93 & 0.36 & \cellcolor{second}{0.27} & 0.22 & \cellcolor{best}{0.21} & 0.17 & \phantom{1}\cellcolor{second}{51.15} \\
    & 5pt \E & $\R,\tvec,f_1, f_2,\lambda_1, \lambda_2$ & GeoCalib - $\lambda_1,\lambda_2,f_1,f_2$ & \cellcolor{best}{23.89} & \phantom{1}\cellcolor{best}{4.38} & \cellcolor{best}{0.51} & \cellcolor{best}{0.25} & \cellcolor{best}{0.10} & 0.21 & \cellcolor{second}{0.15} & 120.04 \\
    & 3pt \E & $\R,\tvec$ & GeoCalib - $\lambda_1, \lambda_2,f_1, f_2,\g_1,\g_2$ & 29.35 & \phantom{1}8.14 & 0.35 & 0.27 & 0.22 & \cellcolor{second}{0.21} & 0.17 & \phantom{1}\cellcolor{best}{42.91} \\
    & 3pt \E & $\R,\tvec,f_1,f_2, \lambda_1, \lambda_2 $ & GeoCalib - $\lambda_1, \lambda_2, f_1, f_2, \g_1, \g_2$ & 29.78 & \phantom{1}\cellcolor{second}{4.90} & \cellcolor{second}{0.48} & 0.29 & 0.11 & 0.23 & \cellcolor{best}{0.15} & 122.40 \\
    \bottomrule
\end{tabular}
}    
    \label{tab:poselib_eth3d_synth_C}
\end{table*}

\begin{table*}[ht]
    \centering
    \caption{\textbf{Natural scenes}: results on all scenes of the \ETH dataset, using Poselib RANSAC for synthetic scenario A - \textit{Wild} (\cf Sec.~\ref{sec:scenarioA}).
    The reported statistics are the same as in Tab.~\ref{tab:poselib_eth3d_synth_A}.}
    \resizebox{\textwidth}{!}{\begin{tabular}{ c | r c c | c c c | c c | c c | c}
    \toprule
    & & & & \multicolumn{7}{c}{Poselib - Prague Parks - Synth A} \\
    \midrule
    & Minimal & Refinement & Sample & AVG $(^\circ)$ $\downarrow$ & MED $(^\circ)$ $\downarrow$ & AUC@10 $\uparrow$ & AVG $\epsilon(\lambda)$ $\downarrow$ & MED $\epsilon(\lambda)$ $\downarrow$  & AVG $\xi(f)$ $\downarrow$ & MED $\xi(f)$ $\downarrow$ & Time (ms) $\downarrow$ \\
    \midrule
    \multirow{25}{*}{\rotatebox[origin=c]{90}{$\lambda_1 = \lambda_2$}}     & 9pt \Fkk & $\R,\tvec,f_1,f_2,\lambda_1,\lambda_2$ & \ding{55} & 21.47 & \phantom{1}4.02 & 0.55 & 0.15 & 0.08 & 0.21 & 0.12 & 359.96 \\
    & 10pt \Fkk & $\R,\tvec,f_1,f_2,\lambda_1,\lambda_2$ & \ding{55} & 15.40 & \phantom{1}3.75 & 0.59 & 0.14 & 0.08 & 0.19 & 0.11 & 116.84 \\
    & 8pt \Fk & $\R,\tvec,f,\lambda$ & \ding{55} & 10.92 & \phantom{1}1.82 & 0.74 & 0.12 & \cellcolor{second}{0.06} & \cellcolor{best}{0.16} & 0.07 & 281.96 \\
    & 9pt \Fk & $\R,\tvec,f,\lambda$ & \ding{55} & 11.60 & \phantom{1}1.80 & 0.72 & 0.13 & 0.06 & \cellcolor{second}{0.19} & 0.07 & 143.65 \\
    & 7pt \F & $\R,\tvec,f_1,f_2$ & $\lambda$ = 0 & 33.25 & 19.66 & 0.15 & 0.87 & 0.89 & 0.89 & 0.66 & 115.99 \\
    & 7pt \F & $\R,\tvec,f_1,f_2,\lambda_1,\lambda_2$ & $\lambda = 0$ & 18.55 & \phantom{1}4.11 & 0.56 & 0.17 & 0.09 & 0.24 & 0.13 & \phantom{1}96.77 \\
    & 7pt \F & $\R,\tvec,f_1,f_2,\lambda_1,\lambda_2$ & $\lambda \in \{0.0, -0.6, -1.2\}$ & 13.29 & \phantom{1}3.41 & 0.60 & 0.13 & 0.08 & 0.22 & 0.12 & 136.28 \\
    & 6pt \Ef & $\R,\tvec,f,\lambda$ & $\lambda = 0$ & 11.48 & \phantom{1}1.82 & 0.71 & 0.13 & 0.07 & 0.25 & 0.07 & 110.16 \\
    & 6pt \Ef & $\R,\tvec,f,\lambda$ & $\lambda \in \{0.0, -0.6, -1.2\}$ & \phantom{1}\cellcolor{best}{7.45} & \phantom{1}\cellcolor{best}{1.58} & \cellcolor{best}{0.76} & \cellcolor{best}{0.10} & \cellcolor{best}{0.06} & 0.24 & \cellcolor{best}{0.07} & 139.55 \\
    & 7pt \F & $\R,\tvec,f_1,f_2$ & GeoCalib - $\lambda$ & 20.48 & \phantom{1}6.82 & 0.39 & 0.34 & 0.19 & 0.38 & 0.22 & \phantom{1}98.05 \\
    & 7pt \F & $\R,\tvec,f_1,f_2,\lambda_1,\lambda_2$ &  GeoCalib - $\lambda$ & 16.12 & \phantom{1}3.68 & 0.59 & 0.13 & 0.08 & 0.21 & 0.12 & \phantom{1}96.62 \\
    & 6pt \Ef & $\R,\tvec,f$ & GeoCalib - $\lambda$ & 13.86 & \phantom{1}2.80 & 0.63 & 0.26 & 0.11 & 0.40 & 0.10 & 115.51 \\
    & 6pt \Ef & $\R,\tvec,f,\lambda$ & GeoCalib - $\lambda$ & \phantom{1}\cellcolor{second}{9.20} & \phantom{1}\cellcolor{second}{1.79} & \cellcolor{second}{0.74} & \cellcolor{second}{0.11} & 0.06 & 0.19 & \cellcolor{second}{0.07} & 107.80 \\
    & 5pt \E & $\R, \tvec$ & GeoCalib - $\lambda,f$ & 24.39 & \phantom{1}8.17 & 0.32 & 0.45 & 0.27 & 0.70 & 0.61 & \phantom{1}\cellcolor{second}{55.96} \\
    & 5pt \E & $\R,\tvec,f,\lambda$ & GeoCalib - $\lambda,f$ & 21.36 & \phantom{1}7.22 & 0.40 & 0.17 & 0.11 & 0.47 & 0.31 & 132.33 \\
    & 3pt \E & $\R,\tvec$ & GeoCalib - $\lambda,f,\g$ & 42.89 & 13.67 & 0.24 & 0.45 & 0.27 & 0.70 & 0.61 & \phantom{1}\cellcolor{best}{52.17} \\
    & 3pt \E & $\R,\tvec,f,\lambda$ & GeoCalib - $\lambda,f,\g$ & 46.57 & 12.18 & 0.28 & 0.26 & 0.15 & 0.58 & 0.43 & 156.39 \\
    \midrule
    \midrule
    \multirow{16}{*}{\rotatebox[origin=c]{90}{$\lambda_1 \neq \lambda_2$}}     & 9pt \Fkk & $\R,\tvec,f_1,f_2,\lambda_1,\lambda_2$ & \ding{55} & 26.04 & \phantom{1}4.50 & 0.50 & 0.15 & \cellcolor{best}{0.08} & 0.23 & 0.14 & 448.49 \\
    & 10pt \Fkk & $\R,\tvec,f_1,f_2,\lambda_1, \lambda_2$ & \ding{55} & \cellcolor{second}{14.76} & \phantom{1}\cellcolor{best}{3.75} & \cellcolor{best}{0.58} & \cellcolor{best}{0.13} & \cellcolor{second}{0.08} & \cellcolor{best}{0.18} & \cellcolor{best}{0.12} & 136.15 \\
    & 7pt \F & $\R,\tvec,f_1, f_2$ & $\lambda_1 = \lambda_2$ = 0 & 42.77 & 29.66 & 0.07 & 0.92 & 0.94 & 0.87 & 0.70 & 168.88 \\
    & 7pt \F & $\R,\tvec,f_1, f_2, \lambda_1, \lambda_2$ & $\lambda_1 = \lambda_2 = 0  $ & 23.48 & \phantom{1}4.54 & 0.50 & 0.20 & 0.09 & 0.23 & 0.14 & \cellcolor{second}{112.51} \\
     & 7pt \F & $\R,\tvec,f_1, f_2, \lambda_1, \lambda_2$ & $\lambda_1, \lambda_2 \in \{0.0, -0.6, -1.2\} $ & \cellcolor{best}{13.73} & \phantom{1}\cellcolor{second}{3.85} & \cellcolor{second}{0.55} & \cellcolor{second}{0.13} & 0.09 & \cellcolor{second}{0.22} & 0.13 & 146.05 \\
    & 7pt \F & $\R,\tvec,f_1, f_2$ & GeoCalib - $\lambda_1, \lambda_2$ & 29.44 & 12.48 & 0.22 & 0.53 & 0.43 & 0.51 & 0.39 & 142.98 \\
    & 7pt \F & $\R,\tvec,f_1,f_2,\lambda_1, \lambda_2$ &  GeoCalib - $\lambda_1, \lambda_2$ & 20.00 & \phantom{1}4.13 & 0.53 & 0.15 & 0.09 & 0.24 & \cellcolor{second}{0.13} & 115.08 \\
    & 5pt \E & $\R, \tvec$ & GeoCalib - $\lambda_1, \lambda_2,f_1, f_2$ & 50.31 & 24.73 & 0.11 & 0.51 & 0.46 & 0.73 & 0.63 & 115.53 \\
    & 5pt \E & $\R,\tvec,f_1, f_2,\lambda_1, \lambda_2$ & GeoCalib - $\lambda_1,\lambda_2,f_1,f_2$ & 44.49 & 12.68 & 0.25 & 0.23 & 0.16 & 0.58 & 0.46 & 180.32 \\
    & 3pt \E & $\R,\tvec$ & GeoCalib - $\lambda_1, \lambda_2,f_1, f_2,\g_1,\g_2$ & 54.64 & 29.88 & 0.09 & 0.51 & 0.46 & 0.73 & 0.63 & \phantom{1}\cellcolor{best}{58.11} \\
    & 3pt \E & $\R,\tvec,f_1,f_2, \lambda_1, \lambda_2 $ & GeoCalib - $\lambda_1, \lambda_2, f_1, f_2, \g_1, \g_2$ & 54.16 & 17.86 & 0.21 & 0.30 & 0.18 & 0.63 & 0.53 & 168.32 \\
    \bottomrule
\end{tabular}
}    
    \label{tab:poselib_pp_synth_A}
\end{table*}

\begin{table*}[ht]
    \centering
    \caption{\textbf{Natural scenes}: results on all scenes of the \PP dataset, using Poselib RANSAC for synthetic scenario B - \textit{Small Distortion} (\cf Sec.~\ref{sec:scenarioB}). 
    The reported statistics are the same as in Tab.~\ref{tab:poselib_eth3d_synth_A}.}
    \resizebox{\textwidth}{!}{\begin{tabular}{ c | r c c | c c c | c c | c c | c}
\toprule
    & & & & \multicolumn{7}{c}{Poselib - Prague Parks - Synth B} \\
    \midrule
    & Minimal & Refinement & Sample & AVG $(^\circ)$ $\downarrow$ & MED $(^\circ)$ $\downarrow$ & AUC@10 $\uparrow$ & AVG $\epsilon(\lambda)$ $\downarrow$ & MED $\epsilon(\lambda)$ $\downarrow$  & AVG $\xi(f)$ $\downarrow$ & MED $\xi(f)$ $\downarrow$ & Time (ms) $\downarrow$ \\
    \midrule
    \multirow{21}{*}{\rotatebox[origin=c]{90}{$\lambda_1 = \lambda_2$}}     & 9pt \Fkk & $\R,\tvec,f_1,f_2,\lambda_1,\lambda_2$ & \ding{55} & 14.10 & \phantom{1}2.78 & 0.65 & 0.08 & 0.05 & 0.18 & 0.09 & 309.20 \\
    & 10pt \Fkk & $\R,\tvec,f_1,f_2,\lambda_1,\lambda_2$ & \ding{55} & 11.27 & \phantom{1}2.64 & 0.66 & 0.08 & 0.05 & 0.17 & 0.08 & 102.14 \\
    & 8pt \Fk & $\R,\tvec,f,\lambda$ & \ding{55} & \phantom{1}9.91 & \phantom{1}1.40 & 0.78 & 0.07 & 0.04 & 0.15 & 0.04 & 247.39 \\
    & 9pt \Fk & $\R,\tvec,f,\lambda$ & \ding{55} & \phantom{1}7.40 & \phantom{1}1.44 & 0.79 & 0.08 & 0.04 & \cellcolor{best}{0.11} & 0.05 & 126.39 \\
    & 7pt \F & $\R,\tvec,f_1,f_2$ & $\lambda$ = 0 & 14.05 & \phantom{1}5.06 & 0.49 & 0.15 & 0.15 & 0.34 & 0.19 & \phantom{1}79.85 \\
    & 7pt \F & $\R,\tvec,f_1,f_2,\lambda_1,\lambda_2$ & $\lambda = 0$ & 12.01 & \phantom{1}2.65 & 0.66 & 0.08 & 0.06 & 0.17 & 0.08 & \phantom{1}80.01 \\
    & 6pt \Ef & $\R,\tvec,f,\lambda$ & $\lambda = 0$ & \phantom{1}\cellcolor{second}{6.18} & \phantom{1}\cellcolor{second}{1.38} & \cellcolor{second}{0.81} & \cellcolor{best}{0.06} & \cellcolor{second}{0.04} & 0.14 & \cellcolor{second}{0.04} & \phantom{1}90.22 \\
    & 7pt \F & $\R,\tvec,f_1,f_2$ & GeoCalib - $\lambda$ & 13.15 & \phantom{1}4.04 & 0.53 & 0.12 & 0.08 & 0.28 & 0.12 & \phantom{1}83.80 \\
    & 7pt \F & $\R,\tvec,f_1,f_2,\lambda_1,\lambda_2$ &  GeoCalib - $\lambda$ & \phantom{1}9.63 & \phantom{1}2.73 & 0.66 & 0.08 & 0.05 & 0.17 & 0.09 & \phantom{1}82.51 \\
    & 6pt \Ef & $\R,\tvec,f$ & GeoCalib - $\lambda$ & \phantom{1}7.06 & \phantom{1}2.14 & 0.71 & 0.12 & 0.08 & 0.19 & 0.08 & \phantom{1}97.37 \\
    & 6pt \Ef & $\R,\tvec,f,\lambda$ & GeoCalib - $\lambda$ & \phantom{1}\cellcolor{best}{5.56} & \phantom{1}\cellcolor{best}{1.35} & \cellcolor{best}{0.81} & \cellcolor{second}{0.06} & \cellcolor{best}{0.04} & \cellcolor{second}{0.13} & \cellcolor{best}{0.04} & \phantom{1}95.32 \\
    & 5pt \E & $\R, \tvec$ & GeoCalib - $\lambda,f$ & 23.73 & \phantom{1}6.98 & 0.40 & 0.23 & 0.15 & 0.50 & 0.42 & \phantom{1}\cellcolor{second}{50.36} \\
    & 5pt \E & $\R,\tvec,f,\lambda$ & GeoCalib - $\lambda,f$ & 21.87 & \phantom{1}3.88 & 0.54 & 0.09 & 0.06 & 0.28 & 0.12 & 119.05 \\
    & 3pt \E & $\R,\tvec$ & GeoCalib - $\lambda,f,\g$ & 39.63 & 11.41 & 0.30 & 0.23 & 0.15 & 0.50 & 0.42 & \phantom{1}\cellcolor{best}{44.65} \\
    & 3pt \E & $\R,\tvec,f,\lambda$ & GeoCalib - $\lambda,f,\g$ & 37.50 & \phantom{1}5.81 & 0.43 & 0.14 & 0.07 & 0.36 & 0.16 & 133.60 \\
    \midrule
    \midrule
    \multirow{14}{*}{\rotatebox[origin=c]{90}{$\lambda_1 \neq \lambda_2$}}     & 9pt \Fkk & $\R,\tvec,f_1,f_2,\lambda_1,\lambda_2$ & \ding{55} & 16.16 & \phantom{1}3.00 & \cellcolor{second}{0.64} & \cellcolor{second}{0.08} & \cellcolor{second}{0.06} & 0.18 & \cellcolor{second}{0.09} & 316.79 \\
    & 10pt \Fkk & $\R,\tvec,f_1,f_2,\lambda_1, \lambda_2$ & \ding{55} & \cellcolor{second}{13.65} & \phantom{1}3.00 & \cellcolor{best}{0.64} & \cellcolor{best}{0.08} & \cellcolor{best}{0.06} & \cellcolor{best}{0.15} & 0.09 & 110.23 \\
    & 7pt \F & $\R,\tvec,f_1, f_2$ & $\lambda_1 = \lambda_2$ = 0 & 20.82 & \phantom{1}8.48 & 0.34 & 0.15 & 0.14 & 0.52 & 0.26 & \phantom{1}93.94 \\
    & 7pt \F & $\R,\tvec,f_1, f_2, \lambda_1, \lambda_2$ & $\lambda_1 = \lambda_2 = 0  $ & \cellcolor{best}{11.80} & \phantom{1}\cellcolor{second}{2.87} & 0.63 & 0.08 & 0.06 & 0.18 & 0.09 & \phantom{1}\cellcolor{second}{88.87} \\
    & 7pt \F & $\R,\tvec,f_1, f_2$ & GeoCalib - $\lambda_1, \lambda_2$ & 23.80 & \phantom{1}8.65 & 0.34 & 0.16 & 0.12 & 0.52 & 0.25 & 106.97 \\
    & 7pt \F & $\R,\tvec,f_1,f_2,\lambda_1, \lambda_2$ &  GeoCalib - $\lambda_1, \lambda_2$ & 14.98 & \phantom{1}\cellcolor{best}{2.84} & 0.64 & 0.09 & 0.06 & \cellcolor{second}{0.16} & \cellcolor{best}{0.08} & \phantom{1}94.53 \\
    & 5pt \E & $\R, \tvec$ & GeoCalib - $\lambda_1, \lambda_2,f_1, f_2$ & 48.20 & 20.63 & 0.16 & 0.29 & 0.23 & 0.59 & 0.42 & \phantom{1}92.20 \\
    & 5pt \E & $\R,\tvec,f_1, f_2,\lambda_1, \lambda_2$ & GeoCalib - $\lambda_1,\lambda_2,f_1,f_2$ & 41.94 & \phantom{1}7.36 & 0.39 & 0.15 & 0.07 & 0.45 & 0.19 & 155.63 \\
    & 3pt \E & $\R,\tvec$ & GeoCalib - $\lambda_1, \lambda_2,f_1, f_2,\g_1,\g_2$ & 49.12 & 23.29 & 0.16 & 0.29 & 0.23 & 0.59 & 0.42 & \phantom{1}\cellcolor{best}{53.14} \\
    & 3pt \E & $\R,\tvec,f_1,f_2, \lambda_1, \lambda_2 $ & GeoCalib - $\lambda_1, \lambda_2, f_1, f_2, \g_1, \g_2$ & 44.63 & \phantom{1}9.57 & 0.36 & 0.20 & 0.09 & 0.51 & 0.24 & 155.06 \\
    \bottomrule
\end{tabular}
}    
    \label{tab:poselib_pp_synth_B}
\end{table*}

\begin{table*}[ht]
    \centering
    \caption{\textbf{Natural scenes}: results on all scenes of the \PP dataset, using Poselib RANSAC for synthetic scenario C - \textit{Visible distortion} (\cf Sec.~\ref{sec:scenarioC}). %
    The reported statistics are the same as in Tab.~\ref{tab:poselib_eth3d_synth_A}.}
    \resizebox{\textwidth}{!}{\begin{tabular}{ c | r c c | c c c | c c | c c | c}
     \toprule
    & & & & \multicolumn{7}{c}{Poselib - Prague Parks - Synth C} \\
    \midrule
    & Minimal & Refinement & Sample & AVG $(^\circ)$ $\downarrow$ & MED $(^\circ)$ $\downarrow$ & AUC@10 $\uparrow$ & AVG $\epsilon(\lambda)$ $\downarrow$ & MED $\epsilon(\lambda)$ $\downarrow$  & AVG $\xi(f)$ $\downarrow$ & MED $\xi(f)$ $\downarrow$ & Time (ms) $\downarrow$ \\
    \midrule
    \multirow{25}{*}{\rotatebox[origin=c]{90}{$\lambda_1 = \lambda_2$}}     & 9pt \Fkk & $\R,\tvec,f_1,f_2,\lambda_1,\lambda_2$ & \ding{55} & 22.52 & \phantom{1}4.29 & 0.53 & 0.17 & 0.11 & 0.23 & 0.14 & 417.26 \\
    & 10pt \Fkk & $\R,\tvec,f_1,f_2,\lambda_1,\lambda_2$ & \ding{55} & 15.08 & \phantom{1}3.62 & 0.59 & 0.15 & 0.09 & \cellcolor{best}{0.19} & 0.12 & 127.20 \\
    & 8pt \Fk & $\R,\tvec,f,\lambda$ & \ding{55} & 12.98 & \phantom{1}1.94 & 0.71 & 0.14 & \cellcolor{best}{0.07} & \cellcolor{second}{0.20} & 0.07 & 259.43 \\
    & 9pt \Fk & $\R,\tvec,f,\lambda$ & \ding{55} & 11.82 & \phantom{1}2.01 & 0.70 & 0.13 & 0.08 & 0.22 & 0.08 & 141.35 \\
    & 7pt \F & $\R,\tvec,f_1,f_2$ & $\lambda$ = 0 & 16.68 & \phantom{1}6.72 & 0.40 & 0.23 & 0.15 & 0.36 & 0.21 & \phantom{1}99.14 \\
    & 7pt \F & $\R,\tvec,f_1,f_2,\lambda_1,\lambda_2$ & $\lambda = -0.9$ & 13.60 & \phantom{1}3.82 & 0.57 & 0.14 & 0.09 & 0.23 & 0.13 & 101.64 \\
    & 7pt \F & $\R,\tvec,f_1,f_2,\lambda_1,\lambda_2$ & $\lambda \in \{-0.6, -0.9, -1.2\}$ & 11.19 & \phantom{1}3.35 & 0.60 & 0.14 & 0.09 & 0.22 & 0.12 & 139.49 \\
    & 6pt \Ef & $\R,\tvec,f,\lambda$ & $\lambda = -0.9$ & \phantom{1}\cellcolor{second}{7.97} & \phantom{1}1.95 & \cellcolor{second}{0.71} & \cellcolor{second}{0.11} & 0.07 & 0.29 & 0.07 & 111.99 \\
    & 6pt \Ef & $\R,\tvec,f,\lambda$ & $\lambda \in \{-0.6, -0.9, -1.2\}$ & \phantom{1}\cellcolor{best}{7.46} & \phantom{1}\cellcolor{second}{1.89} & \cellcolor{best}{0.73} & \cellcolor{best}{0.11} & \cellcolor{second}{0.07} & 0.26 & \cellcolor{second}{0.07} & 140.77 \\
    & 7pt \F & $\R,\tvec,f_1,f_2$ & GeoCalib - $\lambda$ & 22.29 & \phantom{1}8.09 & 0.36 & 0.41 & 0.22 & 0.40 & 0.24 & 104.39 \\
    & 7pt \F & $\R,\tvec,f_1,f_2,\lambda_1,\lambda_2$ &  GeoCalib - $\lambda$ & 16.04 & \phantom{1}3.88 & 0.57 & 0.15 & 0.10 & 0.22 & 0.13 & 104.53 \\
    & 6pt \Ef & $\R,\tvec,f$ & GeoCalib - $\lambda$ & 14.50 & \phantom{1}3.18 & 0.58 & 0.30 & 0.13 & 0.60 & 0.12 & 119.20 \\
    & 6pt \Ef & $\R,\tvec,f,\lambda$ & GeoCalib - $\lambda$ & 11.37 & \phantom{1}\cellcolor{best}{1.84} & 0.71 & 0.12 & 0.08 & 0.34 & \cellcolor{best}{0.07} & 110.80 \\
    & 5pt \E & $\R, \tvec$ & GeoCalib - $\lambda,f$ & 26.75 & 10.06 & 0.29 & 0.52 & 0.38 & 0.74 & 0.68 & \phantom{1}\cellcolor{second}{58.57} \\
    & 5pt \E & $\R,\tvec,f,\lambda$ & GeoCalib - $\lambda,f$ & 23.94 & \phantom{1}7.51 & 0.36 & 0.19 & 0.14 & 0.50 & 0.38 & 143.99 \\
    & 3pt \E & $\R,\tvec$ & GeoCalib - $\lambda,f,\g$ & 44.56 & 14.44 & 0.23 & 0.52 & 0.38 & 0.74 & 0.68 & \phantom{1}\cellcolor{best}{51.83} \\
    & 3pt \E & $\R,\tvec,f,\lambda$ & GeoCalib - $\lambda,f,\g$ & 49.75 & 12.96 & 0.26 & 0.28 & 0.17 & 0.61 & 0.49 & 153.71 \\
    \midrule
    \midrule
    \multirow{16}{*}{\rotatebox[origin=c]{90}{$\lambda_1 \neq \lambda_2$}}     & 9pt \Fkk & $\R,\tvec,f_1,f_2,\lambda_1,\lambda_2$ & \ding{55} & 25.42 & \phantom{1}4.61 & 0.50 & 0.19 & 0.10 & 0.26 & 0.13 & 409.77 \\
    & 10pt \Fkk & $\R,\tvec,f_1,f_2,\lambda_1, \lambda_2$ & \ding{55} & \cellcolor{second}{14.86} & \phantom{1}\cellcolor{best}{3.52} & \cellcolor{best}{0.59} & 0.17 & 0.09 & \cellcolor{best}{0.19} & \cellcolor{best}{0.12} & 129.48 \\
    & 7pt \F & $\R,\tvec,f_1, f_2$ & $\lambda_1 = \lambda_2$ = 0 & 27.09 & 11.64 & 0.25 & 0.37 & 0.28 & 0.47 & 0.35 & 116.40 \\
    & 7pt \F & $\R,\tvec,f_1, f_2, \lambda_1, \lambda_2$ & $\lambda_1 = \lambda_2 = -0.9  $ & 19.66 & \phantom{1}3.88 & 0.54 & 0.16 & 0.09 & 0.22 & 0.12 & 106.36 \\
     & 7pt \F & $\R,\tvec,f_1, f_2, \lambda_1, \lambda_2$ & $\lambda_1, \lambda_2 \in \{-0.6, -0.9, -1.2\} $ & \cellcolor{best}{14.05} & \phantom{1}\cellcolor{second}{3.59} & \cellcolor{second}{0.57} & \cellcolor{best}{0.14} & \cellcolor{second}{0.09} & 0.22 & \cellcolor{second}{0.12} & 141.68 \\
    & 7pt \F & $\R,\tvec,f_1, f_2$ & GeoCalib - $\lambda_1, \lambda_2$ & 31.14 & 13.03 & 0.21 & 0.65 & 0.60 & 0.58 & 0.36 & 129.74 \\
    & 7pt \F & $\R,\tvec,f_1,f_2,\lambda_1, \lambda_2$ &  GeoCalib - $\lambda_1, \lambda_2$ & 23.02 & \phantom{1}3.86 & 0.55 & \cellcolor{second}{0.15} & \cellcolor{best}{0.09} & \cellcolor{second}{0.21} & 0.12 & 107.78 \\
    & 5pt \E & $\R, \tvec$ & GeoCalib - $\lambda_1, \lambda_2,f_1, f_2$ & 48.61 & 25.03 & 0.10 & 0.57 & 0.56 & 0.74 & 0.68 & \phantom{1}\cellcolor{second}{97.32} \\
    & 5pt \E & $\R,\tvec,f_1, f_2,\lambda_1, \lambda_2$ & GeoCalib - $\lambda_1,\lambda_2,f_1,f_2$ & 44.39 & 13.22 & 0.25 & 0.24 & 0.17 & 0.60 & 0.52 & 163.67 \\
    & 3pt \E & $\R,\tvec$ & GeoCalib - $\lambda_1, \lambda_2,f_1, f_2,\g_1,\g_2$ & 56.90 & 32.67 & 0.09 & 0.57 & 0.56 & 0.74 & 0.68 & \phantom{1}\cellcolor{best}{56.10} \\
    & 3pt \E & $\R,\tvec,f_1,f_2, \lambda_1, \lambda_2 $ & GeoCalib - $\lambda_1, \lambda_2, f_1, f_2, \g_1, \g_2$ & 53.21 & 18.16 & 0.19 & 0.34 & 0.24 & 0.62 & 0.56 & 165.57 \\
    \bottomrule

\end{tabular}
}    
    \label{tab:poselib_pp_synth_C}
\end{table*}

\subsection{Prior knowledge about cameras}
\label{sec:scenarios}
For the sampling-based strategy, we can adjust the number $k$ of samples and the sampled values $\M U_i = \{\hat{\lambda}_i^1, \hat{\lambda}_i^2, ..., \hat{\lambda}_i^k\}$ based on %
prior knowledge about the cameras.
We study three different scenarios.

\subsubsection{{Scenario A} -  \textit{Wild}}
\label{sec:scenarioA}
In the first scenario, %
we assume no knowledge about the %
cameras. The cameras can have distortions ranging from small to very high. This scenario represents, \eg, images downloaded from the Internet. To simulate this scenario, we distort images from the \ETH dataset.
For each pair of images, we sample undistortion parameters from a distribution $\mathcal{U}$. 
\newtext{We detect features on synthetically distorted images using SuperPoint~\cite{detone2018superpoint}, and match them using LightGlue~\cite{lindenberger2023lightglue}. The same setup for generating point correspondences given pairs of distorted images applies to Scenario B and C  discussed in Sec.~\ref{sec:scenarioB} and Sec.~\ref{sec:scenarioC}, respectively.} %
We apply either the 
same or different distortions based on the studied setup (\Fk \ or \Fkk). 
We define $\mathcal{U}$ as a piecewise distribution, which is uniform between $-1.5$ and $0$, while its density decreases linearly from $-1.5$ to $-1.8$, reaching half the density of the uniform range. 
This is done to simulate that in practice, undistortion parameters in %
the range $[-1.5, 0]$ are more common than in the range $[-1.8, -1.5]$. 
Thus, %
it is natural to sample undistortion parameters from a wide range of parameters for the sampling-based approach. 
We evaluated the sampling-based solvers with $\M U_1 = \M U_2 = \{0,-0.6,-1.2\}$.
Tab.~\ref{tab:poselib_eth3d_synth_A} shows the results for the \ETH dataset.
The Refinement column in the provided tables %
indicates which parameters are optimized inside LO. 
We show results for cameras with shared intrinsics ($\lambda_1 = \lambda_2)$ %
and for two different cameras ($\lambda_1 \not= \lambda_2)$.
In this scenario, we also tested variants in which we optimize different distortions even for cameras with the same distortion. 
As can be seen in Tab.~\ref{tab:poselib_eth3d_synth_A}, the 6pt \Ef \ solver with the sampling-based strategy with $\M U_1 = \M U_2 = \{0,-0.6,-1.2\}$ outperforms the dedicated minimal radial distortion solvers that are applied in the first step of RANSAC \newtext{and the solvers used in combination with GeoCalib for the case of two equal cameras. For the case of two different cameras the best results are obtained by utilizing GeoCalib predictions with the 5pt \E \ solver.}

\subsubsection{Scenario B  - \textit{Small distortion}}
\label{sec:scenarioB}
In the second scenario, we simulate prior knowledge that our cameras have small distortion, \eg, we are processing images taken by mobile phone or DSLR cameras. 
To simulate this scenario, we distort the feature points with distortions corresponding to  undistortion parameters uniformly sampled from the interval $[-0.3, 0]$. 
In this case, it makes sense to run the 6pt \Ef \ and the 7pt \F \ solver in the sampling-based strategy only once with a small undistortion parameter. We decided to use $\M U_1 = \M U_2 = \{0\}$ to simulate the standard baseline.
Tab.~\ref{tab:poselib_eth3d_synth_B} shows the results for this scenario.
\newtext{For both cases the proposed sampling-based and learning-based prior strategies perform significantly better than the dedicated radial distortion solvers. Considering this scenario with shared intrinsics ($\lambda_1 = \lambda_2$), it can be seen that the baseline 6pt \Ef \ solver used in combination with the sampling strategy or the GeoCalib predictions performs the best. Similar results can be observed for the case of two different cameras ($\lambda_1 \neq \lambda_2$), where the strategy utilizing GeoCalib provides slightly better accuracy. Here we note, that the overall runtime of the sampling strategy is significantly lower since it does not require a costly neural network inference.}%

\begin{table*}[ht!]
    \centering
    \caption{Results for the \ROTUNDA scene using Poselib RANSAC. The reported statistics are the same as in Tab.~\ref{tab:poselib_eth3d_synth_A}.}
    \resizebox{1.0\linewidth}{!}{
    \begin{tabular}{ c | r c c | c c c | c c | c c | c}
    \toprule
    & & & & \multicolumn{7}{c}{Poselib - \ROTUNDA} \\
    \midrule
    & Minimal & Refinement & Sample & AVG $(^\circ)$ $\downarrow$ & MED $(^\circ)$ $\downarrow$ & AUC@10 $\uparrow$ & AVG $\epsilon(\lambda)$ $\downarrow$ & MED $\epsilon(\lambda)$ $\downarrow$  & AVG $\xi(f)$ $\downarrow$ & MED $\xi(f)$ $\downarrow$ & Time (ms) $\downarrow$ \\
    \midrule
    \multirow{25}{*}{\rotatebox[origin=c]{90}{$\lambda_1 = \lambda_2$}}     & 9pt \Fkk & $\R,\tvec,f_1,f_2,\lambda_1,\lambda_2$ & \ding{55} & 26.22 & \phantom{1}8.60 & 0.35 & 0.59 & 0.13 & 0.27 & 0.14 & 3057.36 \\
    & 10pt \Fkk & $\R,\tvec,f_1,f_2,\lambda_1,\lambda_2$ & \ding{55} & 26.56 & \phantom{1}7.68 & 0.37 & 0.58 & 0.12 & 0.25 & 0.14 & \phantom{1}228.54 \\
    & 8pt \Fk & $\R,\tvec,f,\lambda$ & \ding{55} & 25.81 & \phantom{1}4.37 & 0.48 & 0.49 & 0.12 & 0.29 & 0.08 & 1944.54 \\
    & 9pt \Fk & $\R,\tvec,f,\lambda$ & \ding{55} & 26.21 & \phantom{1}6.09 & 0.44 & 0.64 & 0.14 & 0.28 & 0.11 & \phantom{1}495.51 \\
    & 7pt \F & $\R,\tvec,f_1,f_2$ & $\lambda$ = 0 & 26.00 & 11.66 & 0.29 & 0.53 & 0.09 & 0.41 & 0.22 & \phantom{11}89.91 \\
    & 7pt \F & $\R,\tvec,f_1,f_2,\lambda_1,\lambda_2$ & $\lambda = 0$ & 23.58 & \phantom{1}8.45 & 0.35 & 0.41 & 0.12 & 0.30 & 0.16 & \phantom{11}82.09 \\
    & 7pt \F & $\R,\tvec,f_1,f_2,\lambda_1,\lambda_2$ & $\lambda \in \{0.0, -0.6, -1.2\}$ & 20.94 & \phantom{1}6.50 & 0.41 & 0.27 & 0.11 & 0.25 & 0.13 & \phantom{1}271.65 \\
    & 6pt \Ef & $\R,\tvec,f,\lambda$ & $\lambda = 0$ & 19.98 & \phantom{1}3.78 & 0.52 & 0.36 & 0.10 & 0.41 & 0.07 & \phantom{1}171.15 \\
    & 6pt \Ef & $\R,\tvec,f,\lambda$ & $\lambda \in \{0.0, -0.6, -1.2\}$ & \cellcolor{second}{17.80} & \phantom{1}\cellcolor{second}{2.83} & \cellcolor{second}{0.58} & 0.23 & \cellcolor{second}{0.08} & 0.31 & \cellcolor{second}{0.05} & \phantom{1}347.41 \\
    & 7pt \F & $\R,\tvec,f_1,f_2$ & GeoCalib - $\lambda$ & 20.77 & \phantom{1}7.24 & 0.38 & 0.17 & 0.14 & 0.27 & 0.14 & \phantom{11}76.40 \\
    & 7pt \F & $\R,\tvec,f_1,f_2,\lambda_1,\lambda_2$ &  GeoCalib - $\lambda$ & 19.79 & \phantom{1}6.26 & 0.42 & 0.19 & 0.09 & \cellcolor{best}{0.24} & 0.12 & \phantom{11}77.35 \\
    & 6pt \Ef & $\R,\tvec,f$ & GeoCalib - $\lambda$ & 17.94 & \phantom{1}3.47 & 0.55 & \cellcolor{second}{0.16} & 0.14 & 0.55 & 0.07 & \phantom{1}154.75 \\
    & 6pt \Ef & $\R,\tvec,f,\lambda$ & GeoCalib - $\lambda$ & \cellcolor{best}{17.10} & \phantom{1}\cellcolor{best}{2.71} & \cellcolor{best}{0.59} & \cellcolor{best}{0.13} & \cellcolor{best}{0.08} & 0.40 & \cellcolor{best}{0.05} & \phantom{1}151.41 \\
    & 5pt \E & $\R, \tvec$ & GeoCalib - $\lambda,f$ & 31.92 & \phantom{1}8.60 & 0.35 & 0.20 & 0.19 & 0.32 & 0.17 & \phantom{11}\cellcolor{second}{46.38} \\
    & 5pt \E & $\R,\tvec,f,\lambda$ & GeoCalib - $\lambda,f$ & 31.20 & \phantom{1}7.99 & 0.37 & 0.21 & 0.11 & \cellcolor{second}{0.24} & 0.11 & \phantom{1}105.89 \\
    & 3pt \E & $\R,\tvec$ & GeoCalib - $\lambda,f,\g$ & 46.06 & 25.60 & 0.26 & 0.20 & 0.19 & 0.32 & 0.17 & \phantom{11}\cellcolor{best}{35.77} \\
    & 3pt \E & $\R,\tvec,f,\lambda$ & GeoCalib - $\lambda,f,\g$ & 47.74 & 21.25 & 0.26 & 0.47 & 0.25 & 0.28 & 0.14 & \phantom{1}104.24 \\
    \midrule
    \midrule
    \multirow{16}{*}{\rotatebox[origin=c]{90}{$\lambda_1 \neq \lambda_2$}}     & 9pt \Fkk & $\R,\tvec,f_1,f_2,\lambda_1,\lambda_2$ & \ding{55} & 40.40 & 12.64 & 0.29 & 1.01 & 0.25 & 0.37 & 0.25 & 3830.00 \\
    & 10pt \Fkk & $\R,\tvec,f_1,f_2,\lambda_1, \lambda_2$ & \ding{55} & 40.16 & 11.54 & \cellcolor{second}{0.31} & 1.09 & 0.25 & 0.38 & 0.25 & \phantom{1}248.94 \\
    & 7pt \F & $\R,\tvec,f_1, f_2$ & $\lambda_1 = \lambda_2$ = 0 & 39.74 & 16.29 & 0.24 & 0.55 & \cellcolor{best}{0.09} & 0.50 & 0.34 & \phantom{11}82.51 \\
    & 7pt \F & $\R,\tvec,f_1, f_2, \lambda_1, \lambda_2$ & $\lambda_1 = \lambda_2 = 0  $ & 38.22 & 12.58 & 0.29 & 0.51 & 0.22 & 0.39 & 0.29 & \phantom{11}76.45 \\
     & 7pt \F & $\R,\tvec,f_1, f_2, \lambda_1, \lambda_2$ & $\lambda_1, \lambda_2 \in \{0.0, -0.6, -1.2\} $ & \cellcolor{best}{35.43} & \phantom{1}\cellcolor{best}{9.66} & \cellcolor{best}{0.34} & 0.39 & 0.18 & 0.35 & 0.26 & \phantom{1}320.18 \\
    & 7pt \F & $\R,\tvec,f_1, f_2$ & GeoCalib - $\lambda_1, \lambda_2$ & 39.12 & 14.61 & 0.23 & \cellcolor{best}{0.20} & 0.17 & 0.44 & 0.32 & \phantom{11}79.43 \\
    & 7pt \F & $\R,\tvec,f_1,f_2,\lambda_1, \lambda_2$ &  GeoCalib - $\lambda_1, \lambda_2$ & \cellcolor{second}{36.69} & 11.47 & 0.31 & 0.30 & \cellcolor{second}{0.16} & 0.38 & 0.28 & \phantom{11}75.13 \\
    & 5pt \E & $\R, \tvec$ & GeoCalib - $\lambda_1, \lambda_2,f_1, f_2$ & 42.40 & 15.38 & 0.24 & \cellcolor{second}{0.23} & 0.22 & 0.35 & 0.23 & \phantom{11}\cellcolor{second}{59.71} \\
    & 5pt \E & $\R,\tvec,f_1, f_2,\lambda_1, \lambda_2$ & GeoCalib - $\lambda_1,\lambda_2,f_1,f_2$ & 41.65 & \cellcolor{second}{10.80} & 0.30 & 0.27 & 0.17 & \cellcolor{best}{0.29} & \cellcolor{best}{0.18} & \phantom{1}106.16 \\
    & 3pt \E & $\R,\tvec$ & GeoCalib - $\lambda_1, \lambda_2,f_1, f_2,\g_1,\g_2$ & 53.47 & 38.82 & 0.16 & 0.23 & 0.22 & 0.35 & 0.23 & \phantom{11}\cellcolor{best}{28.70} \\
    & 3pt \E & $\R,\tvec,f_1,f_2, \lambda_1, \lambda_2 $ & GeoCalib - $\lambda_1, \lambda_2, f_1, f_2, \g_1, \g_2$ & 52.26 & 33.52 & 0.19 & 0.57 & 0.33 & \cellcolor{second}{0.33} & \cellcolor{second}{0.21} & \phantom{11}87.66 \\
    \bottomrule
       
\end{tabular}

    }
    \label{tab:poselib_rotunda}
\end{table*}

\begin{table*}[ht!]
    \centering
    \caption{Results for the \VITUS scene using Poselib RANSAC. The reported statistics are the same as in Tab.~\ref{tab:poselib_eth3d_synth_A}.}
    \resizebox{1.0\linewidth}{!}{
    \begin{tabular}{ c | r c c | c c c | c c | c c | c}    
       \toprule
    & & & & \multicolumn{7}{c}{Poselib - \CATHEDRAL} \\
    \midrule
    & Minimal & Refinement & Sample & AVG $(^\circ)$ $\downarrow$ & MED $(^\circ)$ $\downarrow$ & AUC@10 $\uparrow$ & AVG $\epsilon(\lambda)$ $\downarrow$ & MED $\epsilon(\lambda)$ $\downarrow$  & AVG $\xi(f)$ $\downarrow$ & MED $\xi(f)$ $\downarrow$ & Time (ms) $\downarrow$ \\
    \midrule
    \multirow{25}{*}{\rotatebox[origin=c]{90}{$\lambda_1 = \lambda_2$}}     & 9pt \Fkk & $\R,\tvec,f_1,f_2,\lambda_1,\lambda_2$ & \ding{55} & 14.98 & 2.81 & 0.60 & 0.27 & 0.07 & 0.25 & 0.09 & 1558.69 \\
    & 10pt \Fkk & $\R,\tvec,f_1,f_2,\lambda_1,\lambda_2$ & \ding{55} & 13.05 & 2.51 & 0.63 & 0.25 & 0.06 & 0.20 & 0.08 & \phantom{1}195.15 \\
    & 8pt \Fk & $\R,\tvec,f,\lambda$ & \ding{55} & 10.16 & 1.41 & 0.73 & 0.23 & 0.05 & 0.19 & 0.05 & \phantom{1}923.78 \\
    & 9pt \Fk & $\R,\tvec,f,\lambda$ & \ding{55} & 10.41 & 1.48 & 0.72 & 0.25 & 0.05 & 0.21 & 0.05 & \phantom{1}318.37 \\
    & 7pt \F & $\R,\tvec,f_1,f_2$ & $\lambda$ = 0 & 21.56 & 9.44 & 0.37 & 0.69 & 0.72 & 0.59 & 0.26 & \phantom{1}149.51 \\
    & 7pt \F & $\R,\tvec,f_1,f_2,\lambda_1,\lambda_2$ & $\lambda = 0$ & 15.04 & 2.84 & 0.60 & 0.24 & 0.07 & 0.25 & 0.09 & \phantom{1}117.09 \\
    & 7pt \F & $\R,\tvec,f_1,f_2,\lambda_1,\lambda_2$ & $\lambda \in \{0.0, -0.6, -1.2\}$ & 11.91 & 2.34 & 0.65 & 0.18 & 0.06 & 0.23 & 0.08 & \phantom{1}239.03 \\
    & 6pt \Ef & $\R,\tvec,f,\lambda$ & $\lambda = 0$ & \phantom{1}9.15 & 1.47 & 0.72 & 0.21 & 0.05 & 0.34 & 0.05 & \phantom{1}161.15 \\
    & 6pt \Ef & $\R,\tvec,f,\lambda$ & $\lambda \in \{0.0, -0.6, -1.2\}$ & \phantom{1}7.24 & \cellcolor{best}{1.23} & \cellcolor{second}{0.77} & 0.15 & \cellcolor{second}{0.04} & 0.25 & 0.04 & \phantom{1}251.30 \\
    & 7pt \F & $\R,\tvec,f_1,f_2$ & GeoCalib - $\lambda$ & 12.49 & 2.71 & 0.62 & 0.14 & 0.06 & 0.26 & 0.09 & \phantom{1}109.99 \\
    & 7pt \F & $\R,\tvec,f_1,f_2,\lambda_1,\lambda_2$ &  GeoCalib - $\lambda$ & 11.68 & 2.32 & 0.65 & 0.16 & 0.06 & 0.22 & 0.08 & \phantom{1}114.35 \\
    & 6pt \Ef & $\R,\tvec,f$ & GeoCalib - $\lambda$ & \phantom{1}7.70 & 1.38 & 0.76 & \cellcolor{best}{0.13} & 0.04 & 0.27 & 0.04 & \phantom{1}149.00 \\
    & 6pt \Ef & $\R,\tvec,f,\lambda$ & GeoCalib - $\lambda$ & \phantom{1}7.18 & \cellcolor{second}{1.23} & \cellcolor{best}{0.77} & \cellcolor{second}{0.13} & \cellcolor{best}{0.04} & 0.23 & 0.04 & \phantom{1}148.29 \\
    & 5pt \E & $\R, \tvec$ & GeoCalib - $\lambda,f$ & \phantom{1}\cellcolor{best}{6.48} & 1.82 & 0.74 & 0.13 & 0.07 & \cellcolor{best}{0.03} & \cellcolor{best}{0.02} & \phantom{11}\cellcolor{second}{49.75} \\
    & 5pt \E & $\R,\tvec,f,\lambda$ & GeoCalib - $\lambda,f$ & \phantom{1}7.58 & 2.01 & 0.71 & 0.16 & 0.05 & 0.15 & 0.06 & \phantom{1}142.24 \\
    & 3pt \E & $\R,\tvec$ & GeoCalib - $\lambda,f,\g$ & \phantom{1}\cellcolor{second}{6.56} & 1.82 & 0.74 & 0.13 & 0.07 & \cellcolor{second}{0.03} & \cellcolor{second}{0.02} & \phantom{11}\cellcolor{best}{37.21} \\
    & 3pt \E & $\R,\tvec,f,\lambda$ & GeoCalib - $\lambda,f,\g$ & \phantom{1}9.21 & 2.13 & 0.69 & 0.17 & 0.06 & 0.16 & 0.07 & \phantom{1}125.65 \\
    \midrule
    \midrule
    \multirow{16}{*}{\rotatebox[origin=c]{90}{$\lambda_1 \neq \lambda_2$}}     & 9pt \Fkk & $\R,\tvec,f_1,f_2,\lambda_1,\lambda_2$ & \ding{55} & 17.19 & \phantom{1}4.24 & 0.51 & 0.32 & 0.12 & 0.32 & 0.27 & 2942.48 \\
    & 10pt \Fkk & $\R,\tvec,f_1,f_2,\lambda_1, \lambda_2$ & \ding{55} & 15.04 & \phantom{1}3.74 & 0.54 & 0.32 & 0.10 & 0.28 & 0.25 & \phantom{1}283.43 \\
    & 7pt \F & $\R,\tvec,f_1, f_2$ & $\lambda_1 = \lambda_2$ = 0 & 27.93 & 14.45 & 0.22 & 0.64 & 0.68 & 0.62 & 0.39 & \phantom{1}184.35 \\
    & 7pt \F & $\R,\tvec,f_1, f_2, \lambda_1, \lambda_2$ & $\lambda_1 = \lambda_2 = 0  $ & 20.18 & \phantom{1}4.60 & 0.49 & 0.29 & 0.12 & 0.34 & 0.29 & \phantom{1}120.52 \\
     & 7pt \F & $\R,\tvec,f_1, f_2, \lambda_1, \lambda_2$ & $\lambda_1, \lambda_2 \in \{0.0, -0.6, -1.2\} $ & 15.89 & \phantom{1}3.58 & 0.55 & 0.22 & 0.09 & 0.32 & 0.28 & \phantom{1}310.38 \\
    & 7pt \F & $\R,\tvec,f_1, f_2$ & GeoCalib - $\lambda_1, \lambda_2$ & 16.74 & \phantom{1}4.53 & 0.50 & 0.15 & \cellcolor{best}{0.08} & 0.34 & 0.28 & \phantom{1}120.81 \\
    & 7pt \F & $\R,\tvec,f_1,f_2,\lambda_1, \lambda_2$ &  GeoCalib - $\lambda_1, \lambda_2$ & 15.76 & \phantom{1}3.53 & 0.56 & 0.20 & \cellcolor{second}{0.09} & 0.30 & 0.27 & \phantom{1}114.55 \\
    & 5pt \E & $\R, \tvec$ & GeoCalib - $\lambda_1, \lambda_2,f_1, f_2$ & \phantom{1}\cellcolor{second}{8.91} & \phantom{1}\cellcolor{second}{3.18} & \cellcolor{second}{0.60} & \cellcolor{best}{0.13} & 0.09 & \cellcolor{best}{0.17} & \cellcolor{best}{0.05} & \phantom{11}\cellcolor{second}{59.71} \\
    & 5pt \E & $\R,\tvec,f_1, f_2,\lambda_1, \lambda_2$ & GeoCalib - $\lambda_1,\lambda_2,f_1,f_2$ & \phantom{1}\cellcolor{best}{8.79} & \phantom{1}\cellcolor{best}{2.98} & \cellcolor{best}{0.61} & 0.20 & 0.09 & 0.24 & 0.20 & \phantom{1}142.15 \\
    & 3pt \E & $\R,\tvec$ & GeoCalib - $\lambda_1, \lambda_2,f_1, f_2,\g_1,\g_2$ & \phantom{1}9.45 & \phantom{1}3.23 & 0.59 & \cellcolor{second}{0.13} & 0.09 & \cellcolor{second}{0.17} & \cellcolor{second}{0.05} & \phantom{11}\cellcolor{best}{32.26} \\
    & 3pt \E & $\R,\tvec,f_1,f_2, \lambda_1, \lambda_2 $ & GeoCalib - $\lambda_1, \lambda_2, f_1, f_2, \g_1, \g_2$ & 10.16 & \phantom{1}3.21 & 0.60 & 0.22 & 0.09 & 0.25 & 0.21 & \phantom{1}109.06 \\
    \bottomrule
   
\end{tabular}

    }
    \label{tab:poselib_cathedral}
\end{table*}

\begin{table*}[ht!]
    \centering
    \caption{Results for the \Euroc dataset  using Poselib RANSAC. The reported statistics are the same as in Tab.~\ref{tab:poselib_eth3d_synth_A}. We do not include statistics for the radial undistortion parameter estimation since GT data is not provided for division model.}
    \resizebox{1.0\linewidth}{!}{
    \begin{tabular}{ c | r c c | c c c | c c | c}
    \toprule
    & & & & \multicolumn{5}{c}{Poselib - \Euroc} \\
    \midrule
    & Minimal & Refinement & Sample & AVG $(^\circ)$ $\downarrow$ & MED $(^\circ)$ $\downarrow$ & AUC@10 $\uparrow$ & AVG $\xi(f)$ $\downarrow$ & MED $\xi(f)$ $\downarrow$  & Time (ms) $\downarrow$ \\
    \midrule

\multirow{25}{*}{\rotatebox[origin=c]{90}{$\lambda_1 = \lambda_2$}}     & 9pt \Fkk & $\R,\tvec,f_1,f_2,\lambda_1,\lambda_2$ & \ding{55} & 45.33 & 10.89 & 0.30  & 0.33 & 0.15 & 151.58 \\
    & 10pt \Fkk & $\R,\tvec,f_1,f_2,\lambda_1,\lambda_2$ & \ding{55} & 37.14 & \phantom{1}7.92 & 0.35  & 0.25 & 0.11 & \phantom{1}56.37 \\
    & 8pt \Fk & $\R,\tvec,f,\lambda$ & \ding{55} & 26.07 & \phantom{1}5.15 & 0.47  & 0.30 & 0.07 & 105.79 \\
    & 9pt \Fk & $\R,\tvec,f,\lambda$ & \ding{55} & 25.64 & \phantom{1}5.28 & 0.46  & 0.30 & 0.08 & \phantom{1}62.81 \\
    & 7pt \F & $\R,\tvec,f$ & $\lambda$ = 0 & 46.69 & 28.37 & 0.05  & 1.29 & 0.98 & \phantom{1}53.27 \\
    & 7pt \F & $\R,\tvec,f_1,f_2,\lambda_1,\lambda_2$ & $\lambda = 0$ & 37.53 & \phantom{1}9.08 & 0.33  & 0.31 & 0.14 & \phantom{1}49.99 \\
    & 7pt \F & $\R,\tvec,f_1,f_2,\lambda_1,\lambda_2$ & $\lambda \in \{0.0, -0.6, -1.2\}$ & 25.98 & \phantom{1}6.63 & 0.39  & \cellcolor{second}{0.24} & 0.10 & \phantom{1}72.19 \\
    & 6pt \Ef & $\R,\tvec,f,\lambda$ & $\lambda = 0$ & 21.82 & \phantom{1}5.21 & 0.47  & 0.38 & 0.08 & \phantom{1}57.97 \\
    & 6pt \Ef & $\R,\tvec,f,\lambda$ & $\lambda \in \{0.0, -0.6, -1.2\}$ & \cellcolor{best}{15.70} & \phantom{1}\cellcolor{best}{4.28} & \cellcolor{best}{0.53}  & 0.26 & \cellcolor{best}{0.06} & \phantom{1}75.72 \\
    & 7pt \F & $\R,\tvec,f$ & GeoCalib - $\lambda$ & 30.34 & 10.48 & 0.28  & 0.39 & 0.19 & \phantom{1}48.01 \\
    & 7pt \F & $\R,\tvec,f_1,f_2,\lambda_1,\lambda_2$ &  GeoCalib - $\lambda$ & 29.09 & \phantom{1}6.95 & 0.38  & 0.25 & 0.11 & \phantom{1}50.89 \\
    & 6pt \Ef & $\R,\tvec,f$ & GeoCalib - $\lambda$ & 18.34 & \phantom{1}5.46 & 0.46  & 0.36 & 0.09 & \phantom{1}55.52 \\
    & 6pt \Ef & $\R,\tvec,f,\lambda$ & GeoCalib - $\lambda$ & \cellcolor{second}{16.62} & \phantom{1}\cellcolor{second}{4.43} & \cellcolor{second}{0.52}  & 0.27 & \cellcolor{second}{0.06} & \phantom{1}57.95 \\
    & 5pt \E & $\R, \tvec$ & GeoCalib - $\lambda,f$ & 17.41 & \phantom{1}5.97 & 0.42  & 0.26 & 0.16 & \phantom{1}\cellcolor{second}{23.89} \\
    & 5pt \E & $\R,\tvec,f,\lambda$ & GeoCalib - $\lambda,f$ & 20.17 & \phantom{1}6.12 & 0.42  & \cellcolor{best}{0.23} & 0.10 & \phantom{1}65.86 \\
    & 3pt \E & $\R,\tvec$ & GeoCalib - $\lambda,f,\g$ & 23.43 & \phantom{1}6.29 & 0.40  & 0.26 & 0.16 & \phantom{1}\cellcolor{best}{22.48} \\
    & 3pt \E & $\R,\tvec,f,\lambda$ & GeoCalib - $\lambda,f,\g$ & 30.59 & \phantom{1}6.78 & 0.38  & 0.25 & 0.11 & \phantom{1}65.34 \\
\bottomrule    
\end{tabular}

    }
    \label{tab:poselib_euroc}
\end{table*}

\subsubsection{Scenario C - \textit{Visible distortion}}
\label{sec:scenarioC}
In the last scenario, we assume that we know that our images have visible (but unknown) distortion. 
To simulate this, we distort the images with distortions corresponding to undistortion parameters uniformly sampled from the interval $[-1.8, -0.5]$.
For the sampling-based strategy, we tested two different variants: (1)  $\M U_1 = \M U_2 = \{-0.9\}$ and  (2) $\M U_1 = \M U_2 = \{-0.6,-0.9,-1.2\}$. The results are shown in Tab.~\ref{tab:poselib_eth3d_synth_C}. 

\newtext{For the case of shared intrinsics ($\lambda_1 = \lambda_2$) the 6pt \Ef \ solver with the sampling strategy performs the best. For the case of two different cameras ($\lambda_1 \neq \lambda_2$) the strategy using a combination of GeoCalib predictions, the 5pt \E \ solver and refinement of intrinsics in LO performs the best. Again, for both cases the proposed sampling-based and learning-based prior strategies perform significantly better than the dedicated radial distortion solvers.
}

\begin{figure*}[ht!]
    \centering

    \begin{tabular}{ccc}
    \multicolumn{3}{c}{\CATHEDRAL ($\lambda_1 = \lambda_2$)} \\
    \includegraphics[height=0.2\textwidth]{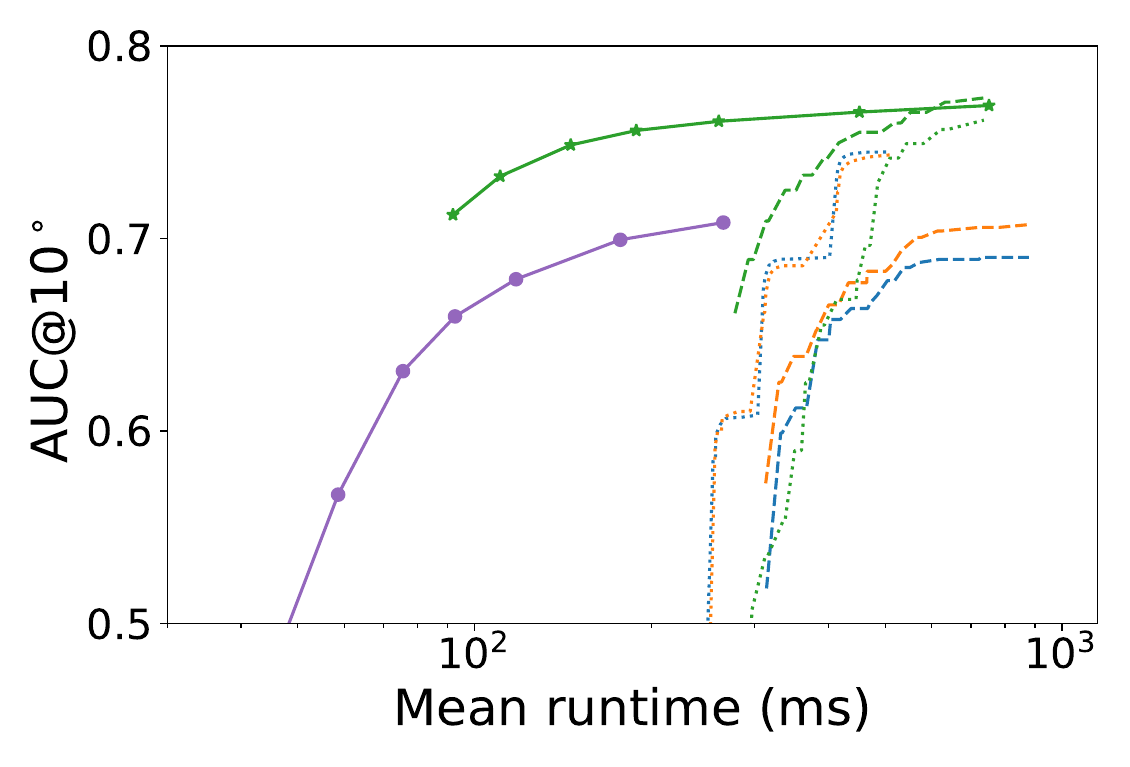} &
    \includegraphics[height=0.2\textwidth]{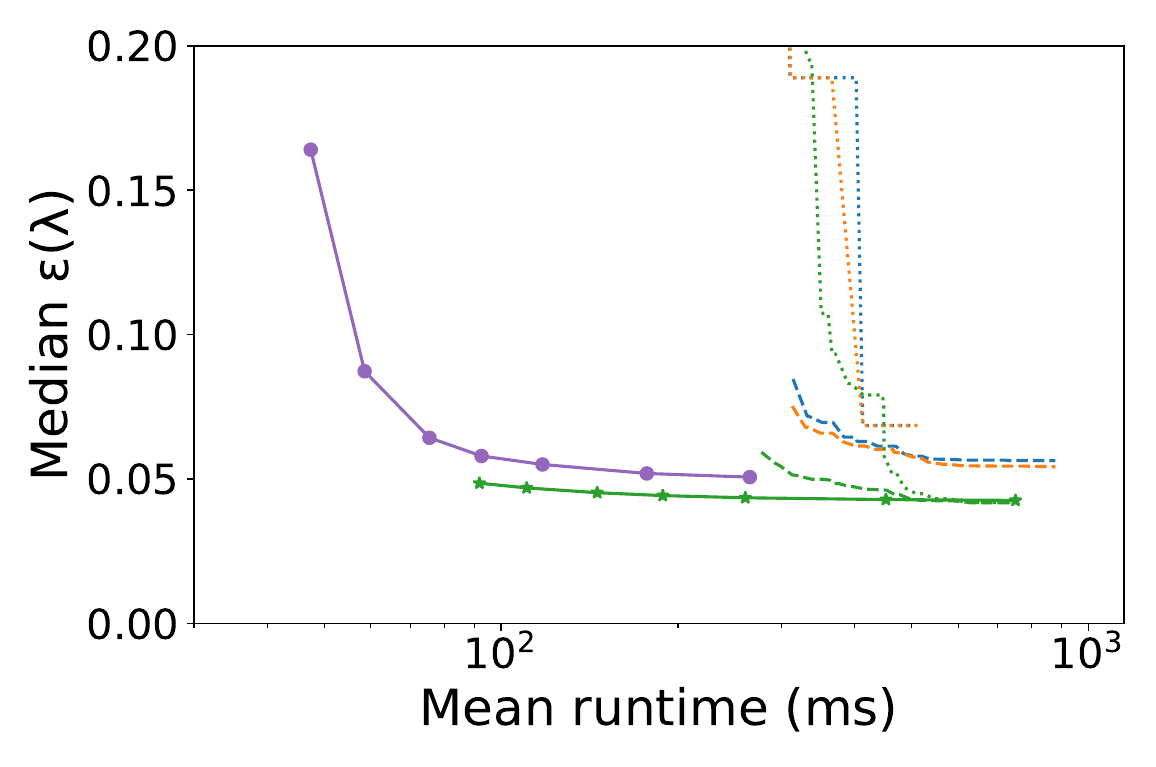} &
    \includegraphics[height=0.2\textwidth]{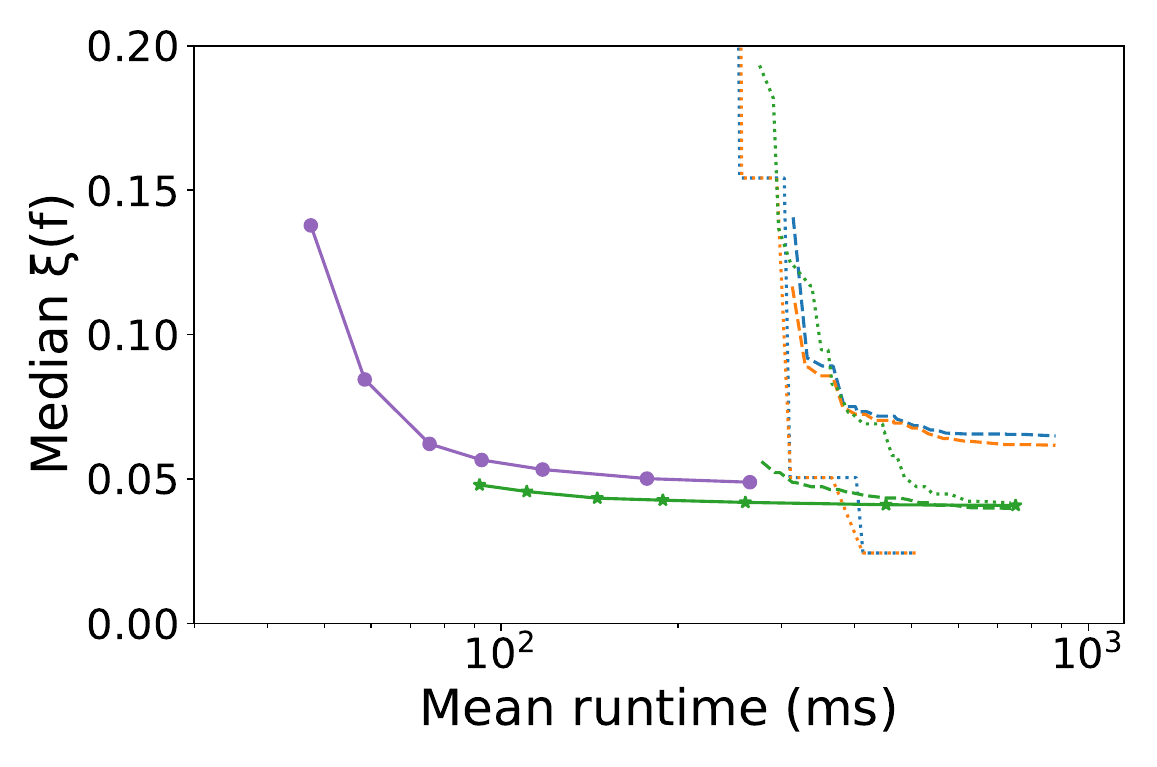} \\
    \multicolumn{3}{c}{\CATHEDRAL ($\lambda_1 \neq \lambda_2$)} \\
    \includegraphics[height=0.2\textwidth]{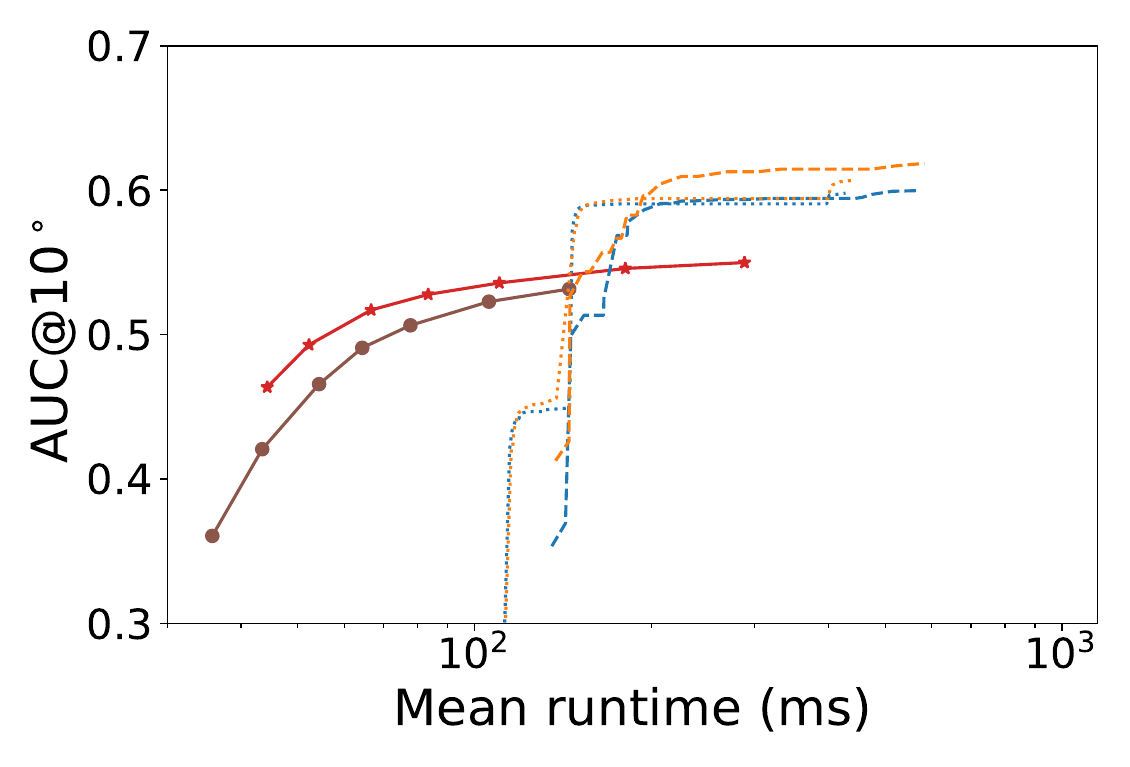} &
    \includegraphics[height=0.2\textwidth]{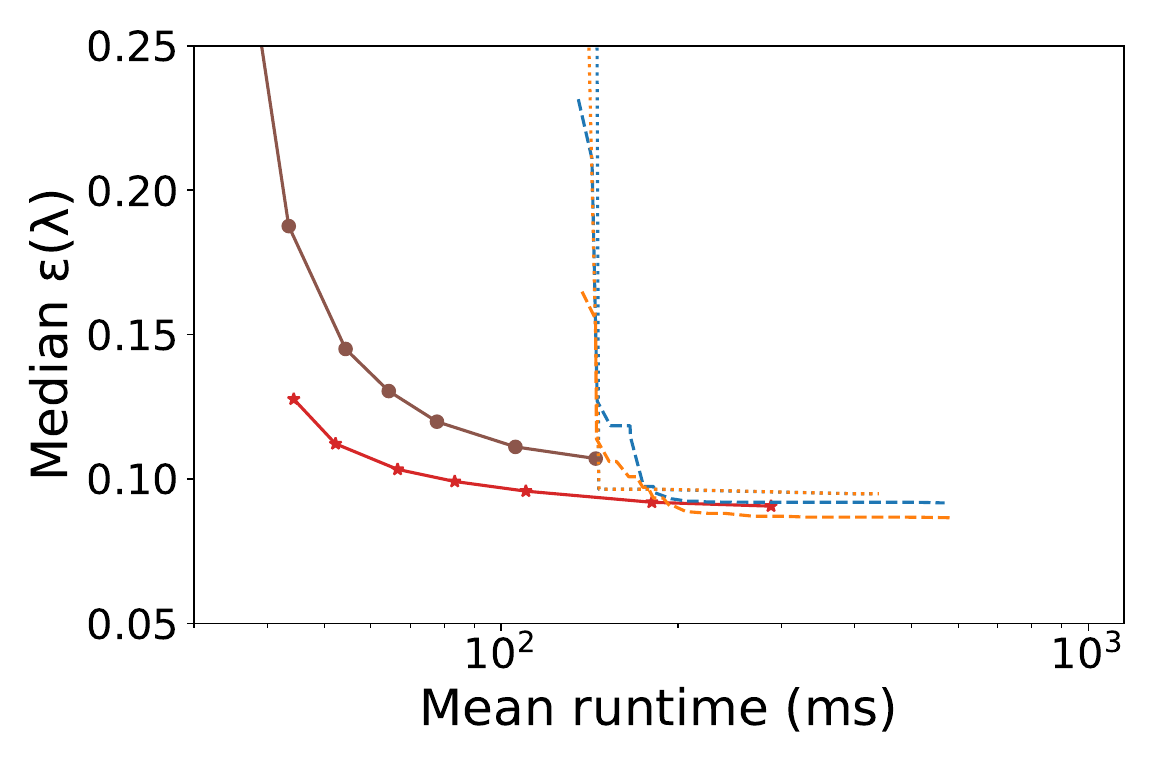} &
    \includegraphics[height=0.2\textwidth]{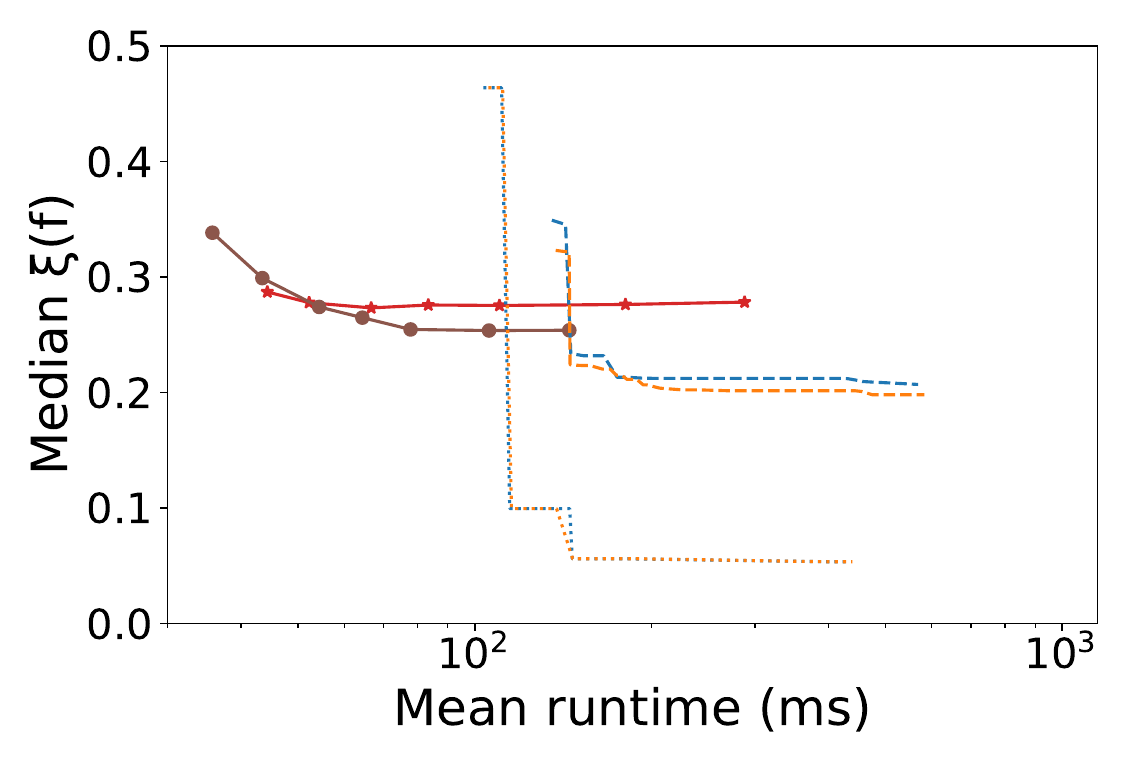} \\

    \multicolumn{3}{c}{\ROTUNDA ($\lambda_1 = \lambda_2$)} \\
    \includegraphics[height=0.2\textwidth]{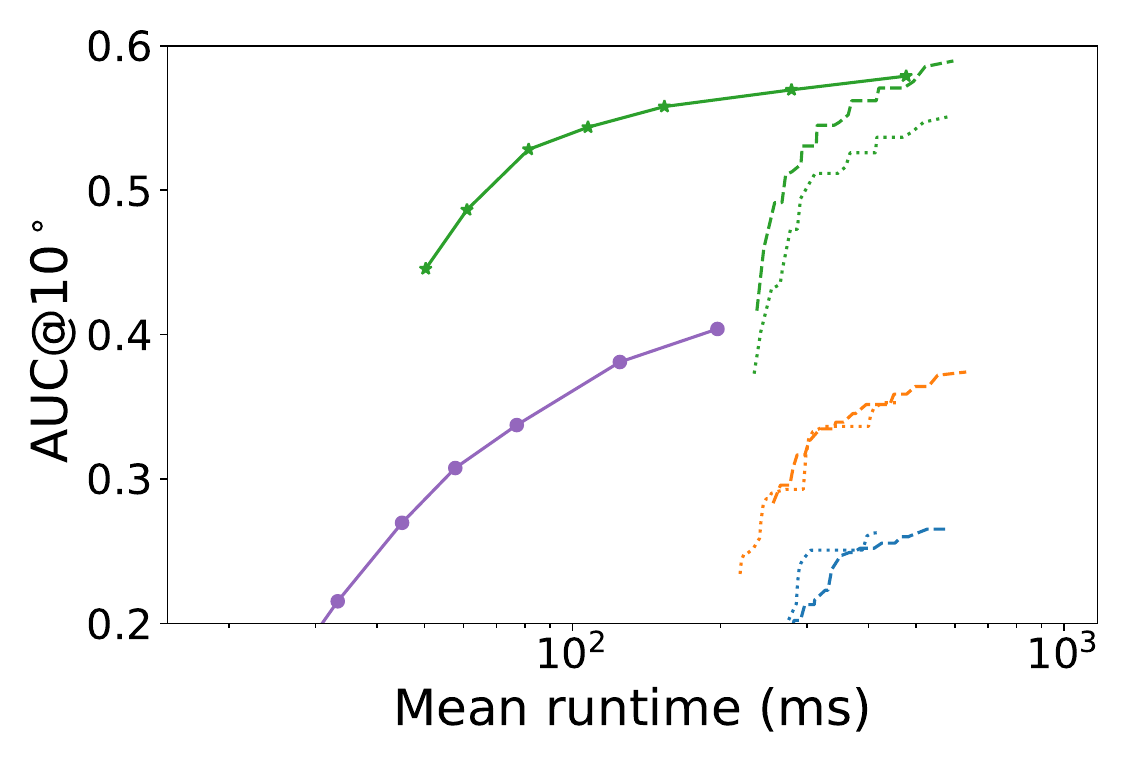} &
    \includegraphics[height=0.2\textwidth]{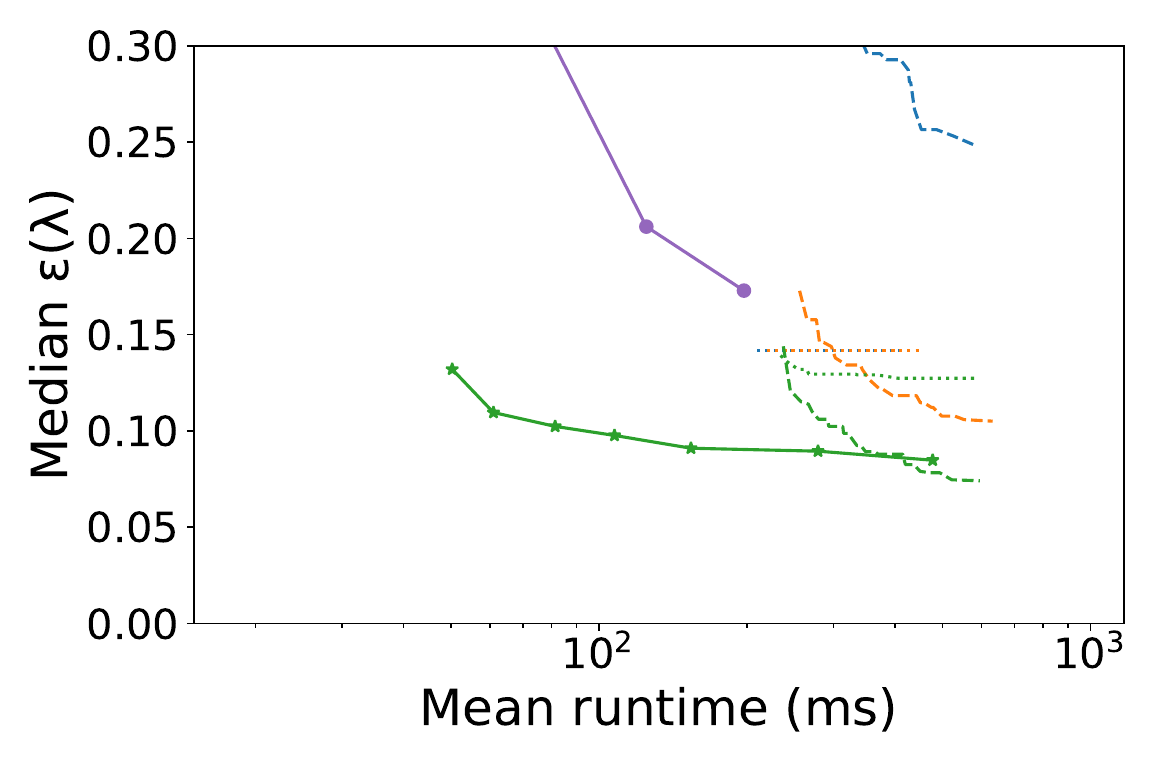} &
    \includegraphics[height=0.2\textwidth]{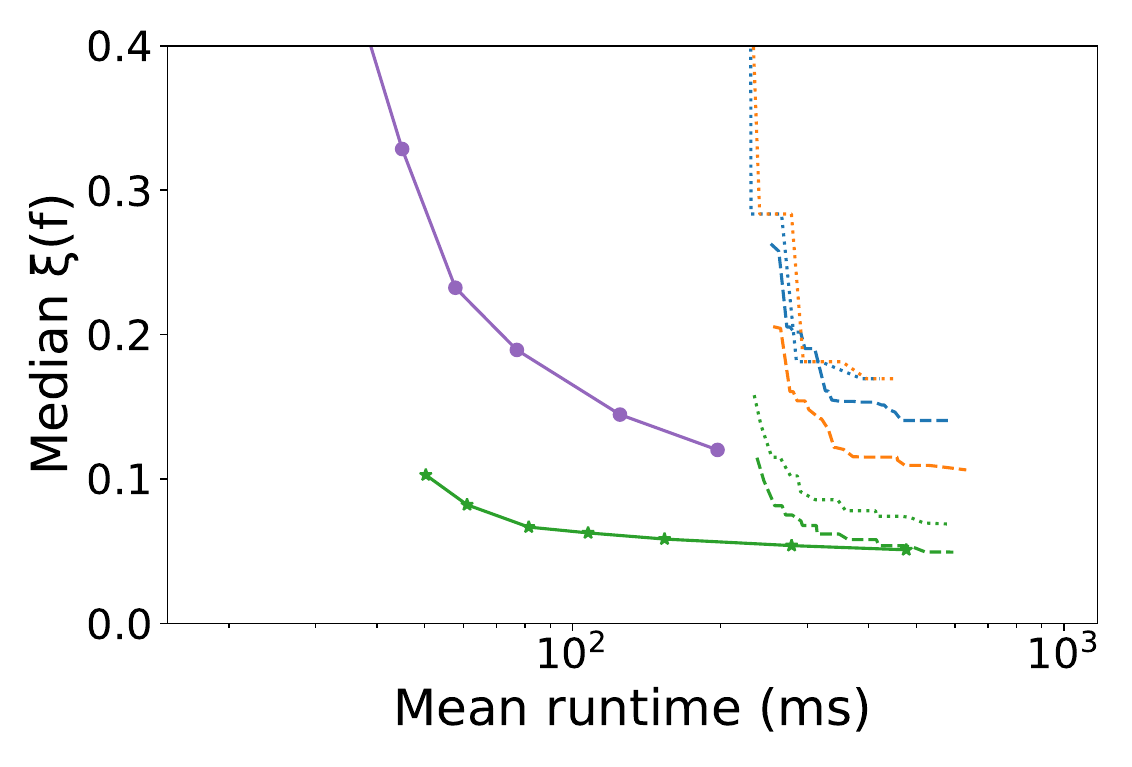} \\
    \multicolumn{3}{c}{\ROTUNDA ($\lambda_1 \neq \lambda_2$)} \\
    \includegraphics[height=0.2\textwidth]{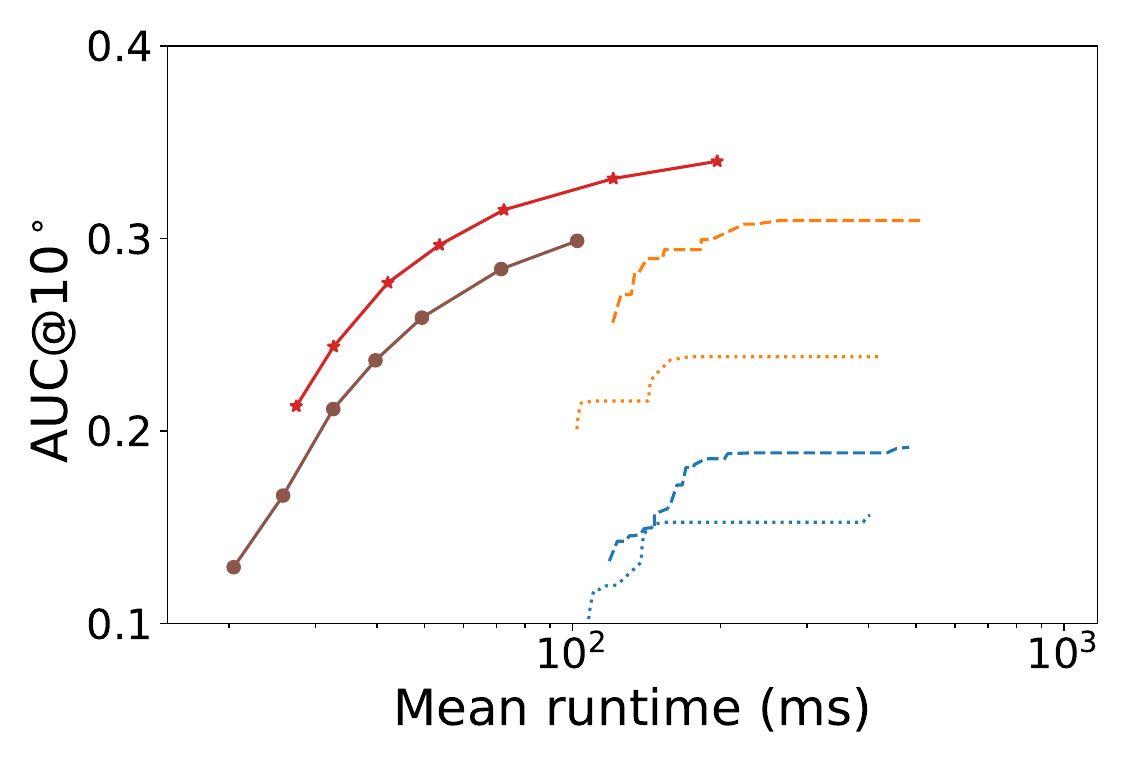} &
    \includegraphics[height=0.2\textwidth]{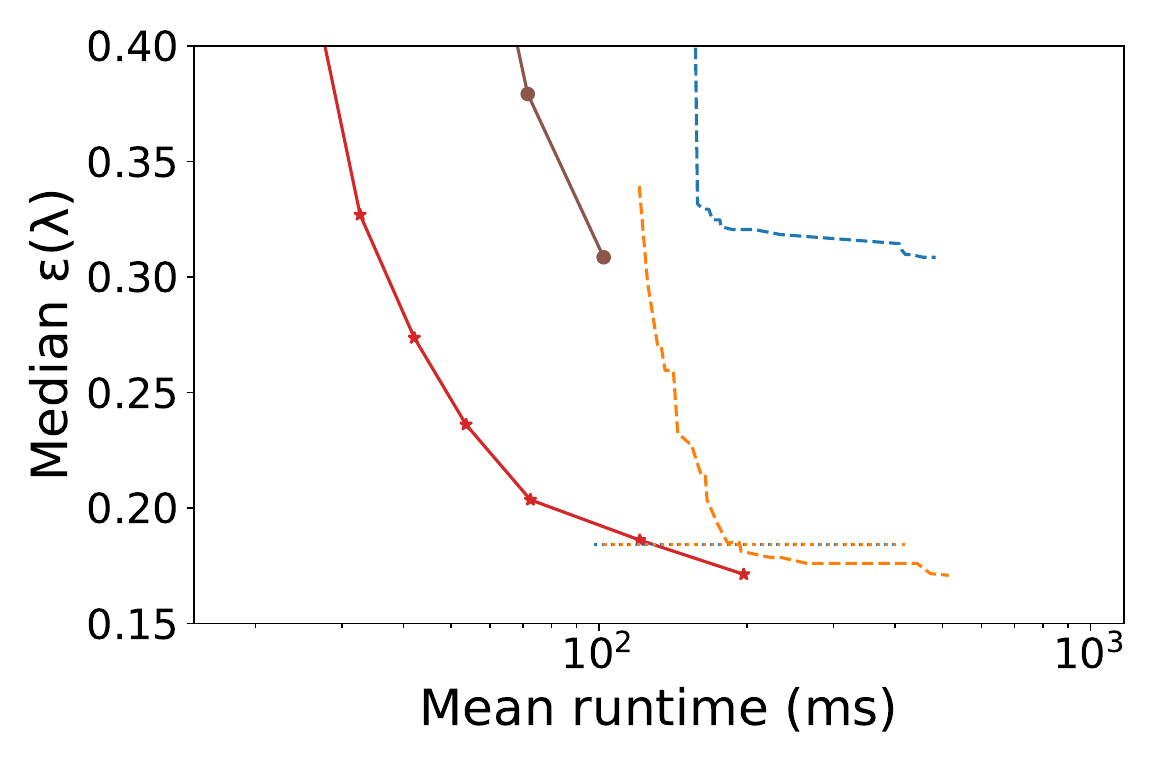} &
    \includegraphics[height=0.2\textwidth]{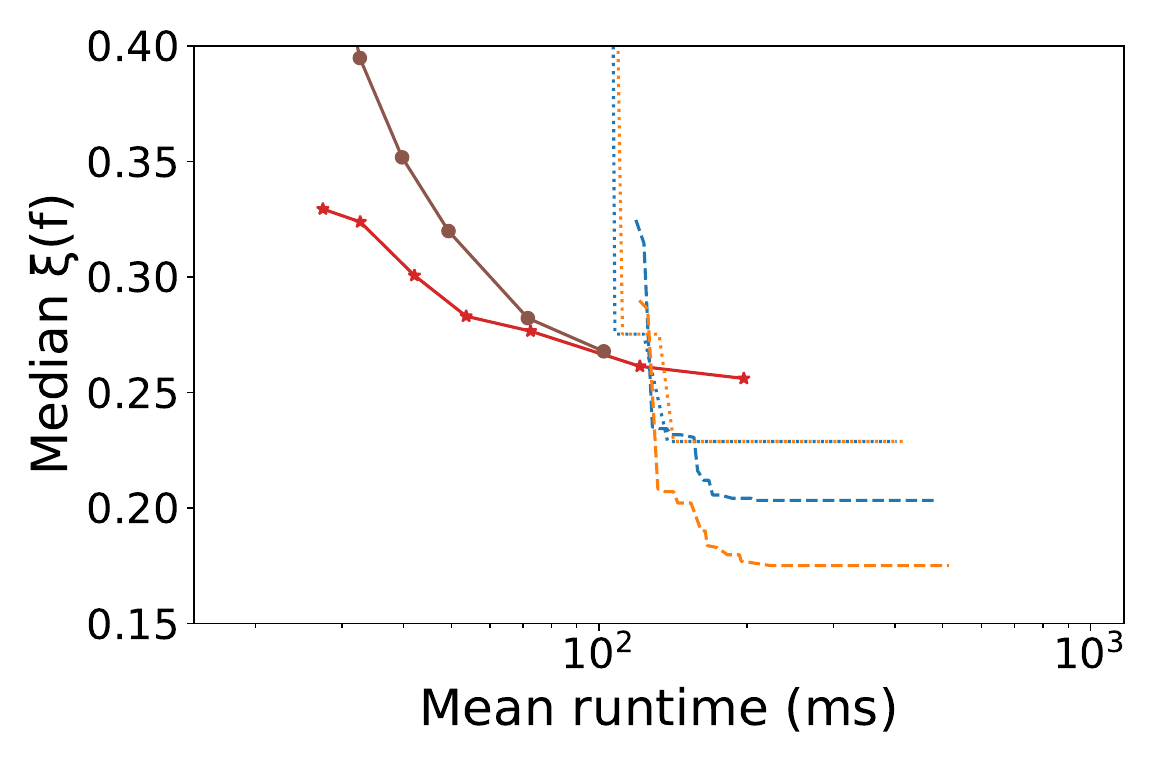} \\
    
    \end{tabular}

    \begin{tikzpicture} 
        \begin{axis}[%
        hide axis, xmin=0,xmax=0,ymin=0,ymax=0,
        legend style={draw=white!15!white, 
        line width = 1pt,
        legend  columns =6, %
        /tikz/every even column/.append style={column sep=0.5cm},
        }
        ]
        
        \addlegendimage{Seaborn1}
        \addlegendentry{\small{3pt \E}};
        \addlegendimage{Seaborn2}
        \addlegendentry{\small{5pt \E}};
        \addlegendimage{Seaborn3}
        \addlegendentry{\small{6pt \Ef}};
        \addlegendimage{Seaborn4}
        \addlegendentry{\small{7pt \F}};
        \addlegendimage{Seaborn5}
        \addlegendentry{\small{9pt \Fk}};   
        \addlegendimage{Seaborn6}
        \addlegendentry{\small{10pt \Fkk}};  
        \end{axis}
    \end{tikzpicture}
    \begin{tikzpicture} 
        \begin{axis}[%
        legend style={cells={align=center}},
        hide axis, xmin=0,xmax=0,ymin=0,ymax=0,
        legend style={draw=white!15!white, 
        line width = 1pt,
        legend  columns =6, %
        /tikz/every even column/.append style={column sep=0.5cm},
        }
        ]

        \addlegendimage{black!30,mark=oplus*}
        \addlegendentry{Dedicated Solvers}

        \addlegendimage{black!30,mark=star}
        \addlegendentry{Sampling  \\ \scriptsize{$\lambda \in \{0,-0.6,-1.2\}$}}
        
        \addlegendimage{black!30,dash pattern=on 1pt off 1pt on 1pt off 1pt}        
        \addlegendentry{GeoCalib \\ \scriptsize{only pose refined}};        
        
        \addlegendimage{black!30,dash pattern=on 2pt off 2pt on 2pt off 2pt}
        \addlegendentry{GeoCalib \\ \scriptsize{pose + intrinsics refined}};                
        \end{axis}
    \end{tikzpicture}
                 
    \caption{Pose AUC@10$^\circ$, mean absolute $\lambda$ errors and relative focal length errors plotted for different total runtimes of the compared methods. For all methods we vary the total number of RANSAC iterations ($\{10, 20, 50, 100, 200, 500, 1000\}$).
    For the methods utilizing the learning-based prior strategy with GeoCalib we also vary the total number of LM iterations to produce the final estimate ($\{1, 2, 5, 30\}$). To plot the curves we always take the best performing setting achieving equal or shorter runtime.}    
    \label{fig:poselib_vitus_time}
\end{figure*}

\subsubsection{Natural Scenes}
\newtext{We repeat the three scenarios, but instead of the \ETH dataset we use the \PP dataset. The results are shown in Tab.~\ref{tab:poselib_pp_synth_A}-\ref{tab:poselib_pp_synth_C}. 
This dataset contains natural scenes. For such scenes the GeoCalib network produces worse intrinsics estimates leading to a worse overall performance of the learning-based prior strategy, especially when visible distortion is present, \ie in scenarios A and C. We note that for the case of shared intrinsics the 6pt \Ef \ solver with the sampling strategy leads to the best results. While for the case of two different cameras the dedicated non-minimal 10pt \Fkk~solver performs the best with the sampling strategy combined with the 7pt \F \ solver achieving slightly worse results. 
}

These experiments on both \ETH and \PP datasets show some important observations: 
(1) The sampling-based and \newtext{learning-based prior strategies} perform similar to or even better than the dedicated minimal radial distortion solvers.
(2) Having additional knowledge about the cameras (even vague knowledge, \eg, that the cameras have visible distortion) can improve the performance of the sampling-based strategy. %
\newtext{(3) The sampling-based strategy with the 6pt \Ef \ solver outperforms the learning-based prior strategy when the cameras can be assumed to have the same intrinsics ($\lambda_1 = \lambda_2$). For the case of two different cameras ($\lambda_1 \neq \lambda_2$), using the GeoCalib predictions may lead to significantly better results than relying on a simple sampling strategy when considering images of man-made structures.
(4) In general, the sampling-
based approach is more robust than the one using
learning-based priors and works well for all tested
datasets.
}

\subsection{Real-World Scenario} 
In the previous experiments, we synthesized distortions to be able to precisely measure the behavior of the different approaches under varying levels of distortion. 
To evaluate the performance of the tested methods under real-world conditions and for cameras with different distortions, 
we use the \Euroc dataset and our own dataset consisting of the \ROTUNDA and \VITUS scenes. 
\newtext{Results are shown in Tab.~\ref{tab:poselib_rotunda} for \ROTUNDA, Tab.~\ref{tab:poselib_cathedral} for \CATHEDRAL and Tab.~\ref{tab:poselib_euroc} for \Euroc. }
These results show that the sampling-based and learning-based prior strategies significantly outperform the dedicated solvers. For the equal camera case ($\lambda_1 = \lambda_2$) the 6pt \Ef \ solver with LO of intrinsics achieves best results when used with both sampling $\M U_1 = \M U_2 = \{0,-0.6,-1.2\}$ and GeoCalib for all scenes. We note that for both of these approaches the resulting accuracy is very similar. The sampling strategy results in increased RANSAC time, but this increase is lower than the time required for GeoCalib inference (see Sec.~\ref{sec:speed_acc}). 

For the case of two different cameras ($\lambda_1 \neq \lambda_2$), the learning-based prior strategy performs the best for the \VITUS scene while 
the sampling-based strategy with the 7pt \F~solver and $\M U_1 = \M U_2 = \{0,-0.6,-1.2\}$ provides the best pose estimates for the \ROTUNDA scene. 
We note that all methods perform worse on the \ROTUNDA scene 
in terms of both the estimated poses and intrinsics suggesting that this scene is significantly more difficult than \VITUS. 
This can also be a reason why GeoCalib's intrinsics estimates are less accurate for \ROTUNDA than for \VITUS, resulting in worse performance of the learning-based prior strategy compared to the sampling-based strategy in the \ROTUNDA different distortion scenario.

We also note that the GeoCalib predictions of focal lengths for the \CATHEDRAL scene are significantly better than relying on the geometry obtained from point correspondences via solvers. Similarly, when GeoCalib is used with refinement of the intrinsics in LO, the focal length error increases significantly, while the pose estimate becomes more accurate. This may be caused by degenerate camera configurations which introduce ambiguity into focal length estimation \cite{bougnoux1998projective}. 
In such configurations, the single-view predictions of intrinsics based on learned visual features and geometric cues may be more accurate than relying on point correspondences and epipolar geometry.

\subsubsection{Speed-accuracy trade-off}
\label{sec:speed_acc}
In some situations, the time for relative pose estimation is limited. To evaluate the viability of the different methods in such scenario we conduct an experiment on both scenes to assess each solver's performance, in terms of the AUC@10$^\circ$ of pose errors, and the median absolute error of the estimated undistortion parameter(s) and \newtext{focal length(s)}, for different numbers of RANSAC iterations. \newtext{For the learning-based prior strategy we also varied the total number of LM iterations to obtain the final intrinsics estimate. We evaluate the runtimes using a 2 GHz Intel Xeon Gold 6338 CPU for RANSAC and an A100 GPU for GeoCalib inference.} The plots of the measured metrics vs. the average run-time are reported in Fig.~\ref{fig:poselib_vitus_time}.
\newtext{The plots show that for the case of shared intrinsics ($\lambda_1 = \lambda_2$) the 6pt \Ef \ solver combined with the sampling strategy provides the best speed-accuracy tradeoff across the board. For the case of two different cameras ($\lambda_1 \neq \lambda_2$) the 7pt solver with sampling performs best in situations low time budget ($<$100 ms). When more time is available for computation the learning-based prior strategy may be more optimal.}

\section{Conclusion}
Modeling radial distortion during relative pose estimation %
is important. %
Yet, (minimal) radial distortion solvers %
are significantly more complex than %
solvers for pinhole cameras, %
in terms of both %
runtime and %
implementation efforts. 
This paper thus asks the question whether minimal radial distortion solvers are actually necessary in practice. 
\newnewtext{To answer this question, we considered two approaches that do not require minimal radial distortion solvers: 
The first samples radial distortion parameters from a fixed set of values rather than estimating them, while the second uses a neural network to predict the parameters. Both approaches uses the sampled / predicted parameters in combination with a pinhole relative pose solver.} 
Extensive experiments 
show that \newnewtext{both of these simple strategies} %
outperforms existing %
distortion solvers. 
\newnewtext{Both approaches are } 
easy to implement and faster than the best-performing minimal distortion solvers. Moreover, on real data, they result in better accuracy than the dedicated radial distortion solvers. 
We conclude that 
the dedicated distortion solvers are not truly necessary in practice. 
\newnewtext{We also show that the sampling-based approach is more robust than the one using learning-based priors and works well for all tested datasets, despite not requiring a GPU.}

\section*{Acknowledgements}

V.K. and Z.B.H. were supported by the NextGenerationEU through the Recovery and Resilience Plan for Slovakia under the project ``InnovAIte Slovakia, Illuminating Pathways for AI-Driven Breakthroughs'' No.~09I02-03-V01-00029.
C. T., Y.D. and Z.K. were supported by the Czech Science Foundation (GAČR) JUNIOR STAR Grant (No. 22-23183M). 
T. S. was supported by the EU Horizon 2020 project RICAIP (grant agreement No. 857306).
Part of the research results was obtained using the computational resources procured in the national project National competence centre for high performance computing (project code: 311070AKF2) funded by European Regional Development Fund, EU Structural Funds Informatization of society, Operational Program Integrated Infrastructure. 

\section*{Data Availability}

The code for the methods and evaluation, along with the new benchmark dataset are available at https://github.com/kocurvik/rdnet.

\bibliography{sn-bibliography}

\end{document}